\documentclass[letterpaper]{article} 
\usepackage{aaai24}  
\usepackage{times}  
\usepackage{helvet}  
\usepackage{courier}  
\usepackage[hyphens]{url}  
\usepackage{graphicx} 
\def\UrlFont{\rm}  
\usepackage{natbib}  
\usepackage{caption} 
\usepackage{booktabs}
\usepackage{multirow}
\usepackage{algorithm}
\usepackage{algorithmic}
\usepackage{amsmath,amsthm,amsfonts,amssymb,amscd}
\usepackage{xcolor}

\theoremstyle{plain}

\theoremstyle{definition}

\theoremstyle{remark}

\newcommand{\eqnref}[1]{Eqn.~(\ref{#1})}
\newcommand{\MYCOMMENT}[1]{\hfill \textit{// #1}}
\usepackage{newfloat}
\usepackage{listings}
\DeclareCaptionStyle{ruled}{labelfont=normalfont,labelsep=colon,strut=off} 
\floatstyle{ruled}
\newfloat{listing}{tb}{lst}{}
\floatname{listing}{Listing}
\title{Stitching Sub-Trajectories with Conditional Diffusion Model \\ for Goal-Conditioned Offline RL}
\author{
    Sungyoon Kim, 
    Yunseon Choi, 
    Daiki E. Matsunaga, 
    Kee-Eung Kim
}
\affiliations{
    Kim Jaechul Graduate School of AI, KAIST\\
    \{sykim, yschoi, dematsunaga\}@ai.kaist.ac.kr, kekim@kaist.ac.kr

}

\usepackage{bibentry}

\begin{document}

\maketitle

\begin{abstract}
Offline Goal-Conditioned Reinforcement Learning (Offline GCRL) is an important problem in RL that focuses on acquiring diverse goal-oriented skills solely from pre-collected behavior datasets. 
In this setting, the reward feedback is typically absent except when the goal is achieved, 
which makes it difficult to learn policies especially from a finite dataset of suboptimal behaviors.
In addition, realistic scenarios involve long-horizon planning, which necessitates the 
extraction of useful skills within sub-trajectories.
Recently, the conditional diffusion model has been shown to be a promising approach to 
generate high-quality long-horizon plans for RL. 
However, their
practicality for the goal-conditioned setting is still limited due to a number of technical assumptions
made by the methods.
In this paper, we propose SSD (Sub-trajectory Stitching with Diffusion), 
a model-based offline GCRL method that leverages the conditional diffusion model to address these limitations. 
In summary, we use the diffusion model that generates future plans conditioned on the target goal and value, 
with the target value estimated from the goal-relabeled offline dataset.
We report state-of-the-art performance in the standard benchmark set of GCRL tasks, and 
demonstrate the capability to successfully stitch the segments of suboptimal trajectories in the offline data to generate high-quality plans.
\end{abstract}

\section{Introduction}
\label{sec:intro}

\begin{figure}[t]
\centering
\includegraphics[width=0.5\textwidth]{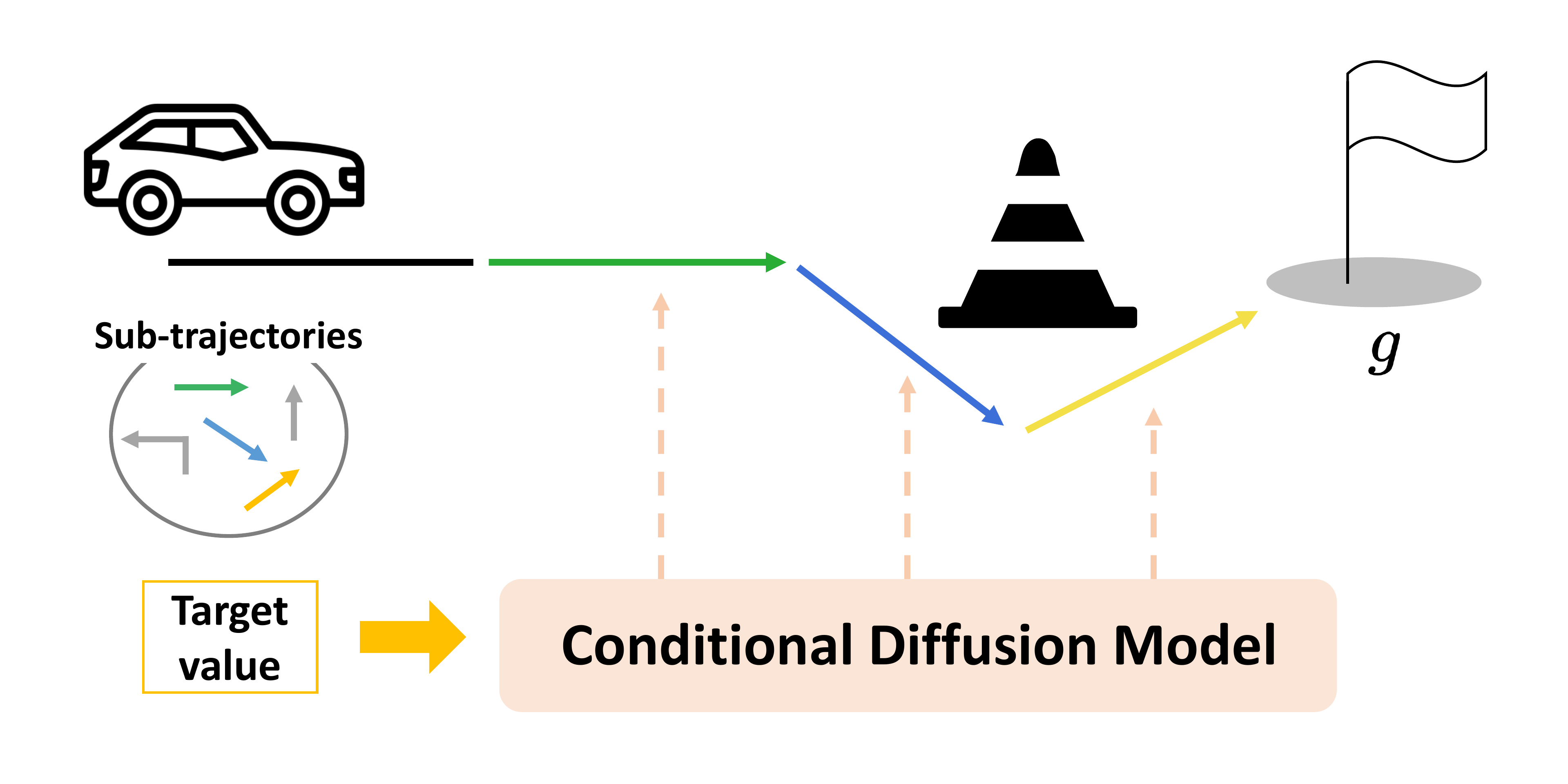} 
\caption{Our approach uses a diffusion model conditioned on the goal and return-to-go that generates sub-trajectories. 
These sub-trajectories are then be stitched together to achieve the goal. 
}
\label{fig:car}
\end{figure}

Offline Reinforcement Learning (Offline RL)~\cite{lange2012, levine2020offline}, which aims to train agents from pre-collected datasets without any further interaction with the environment, has emerged as a promising paradigm in RL, in that it addresses 
the safety issue associated with exploring directly in the environment. One of the most important factors towards successfully training the agent
is to \emph{stitch} together the segments of sub-optimal behaviors in the dataset to generate high-quality behaviors~\cite{kumar2022should}.

This stitching behavior becomes even more important when we consider Goal-Conditioned Reinforcement Learning (GCRL), where the agent aims to learn a policy that achieves specified goals. 
In contrast to traditional RL where the policy is solely conditioned on states, a GCRL policy is additionally conditioned on goals. 
Thus, a GCRL agent is tasked with learning a goal-conditioned policy prepared for diverse goals. 
In order to do so, it is essential
to be able to stitch valuable parts of the behavior in the dataset to generate plans that constitute a policy that achieves a
diverse set of goals.
However, especially with sub-optimal behaviors in the dataset, the sparse nature of the rewards makes it challenging to 
accurately capture the valuable parts and integrate them appropriately.

The challenge originating from the sparse rewards becomes more pronounced when agents are tasked with long planning horizons, 
exemplified by many practical scenarios such as a service robot that needs to carry out a long sequence of complex manipulations to accomplish the goal. 
While current deep RL algorithms excel at learning policies for short-horizon tasks, they struggle to effectively reason over extended planning horizons~\cite{shah2021value}. 
Recently, diffusion-based planning methods have been introduced as a promising approach to generate high-quality long-horizon 
plans~\cite{janner2022planning, ajay2022conditional}.
However, some of the technical assumptions made by the methods make them hard to be used directly for GCRL. 
For example, Diffuser~\cite{janner2022planning} relies on inpainting, where the last state
in the trajectory is set to the goal in order to generate behaviors that terminate at the goal
location. If the length of the plan is conservatively set to be longer than the length of the optimal plan, 
it would end up generating excessively long trajectories in the aforementioned tasks.



In this paper, we propose a novel approach named \textit{SSD (Sub-trajectory Stitching with Diffusion)} that utilizes 
the conditional diffusion model to promote behavior stitching for Offline GCRL, as illustrated in Figure~\ref{fig:car}.
We begin with training the action-value function using multi-step goal chaining with relabeled goals, 
which extends the method in \cite{chebotar2021actionable} to further enrich reward signals.
We then use the estimated action value for conditioning the diffusion model with the \emph{target value}, circumventing the need for 
optimal plan length~\cite{janner2022planning},
explicit subgoals~\cite{florensa2018automatic, chane2021goal} or hierarchical architectures~\cite{li2023hierarchical}.
Regarding the conditional diffusion model, we propose an effective architecture for generating realistic trajectories, named Condition-Prompted-Unet. 
By incorporating transformer blocks instead of convolutional layers, we were able to improve conformity with the dynamics of the environment.
Lastly, we empirically substantiate our approach and demonstrate state-of-the-art performance on Maze2D and Multi2D, where we also qualitatively showcase the stitching capabilities of SSD.


\section{Preliminaries and Related Works} \label{sec:prel_related}
GCRL is formulated as goal-augmented MDP (GA-MDP)~\cite{liu2022goalconditioned}, 
represented by the tuple $(\mathcal{S}, \mathcal{A}, P, R, \mathcal{G}, p_g, \psi, \mu_0, \gamma)$, 
where $\mathcal{S}$ is a set of states, 
$\mathcal{A}$ is a set of actions, 
$P : \mathcal{S} \times \mathcal{A} \rightarrow \Delta(\mathcal{S})$ is the probabilistic state transition function,
$R : \mathcal{S} \times \mathcal{A} \times \mathcal{G} \to \mathbb{R}$ is the reward function,
$\mathcal{G}$ is the set of goals,
$p_g$ is the probability distribution of the desired goal,
$\psi: \mathcal{S} \to \mathcal{G}$ is the state-to-goal mapping,
$\mu_0 \in \Delta(\mathcal{S})$ is the initial state distribution, and 
$\gamma \in [0,1]$ is the discount factor.
We consider a sparse binary reward function 
$R(s_t,a_t,g)\equiv \mathbf{1}(\psi(s_{t})=g)$ which yields a reward of one if and only if the current state satisfies
the goal. 

Given a GA-MDP, we aim to find goal conditioned policy $\pi: \mathcal{S} \times \mathcal{G} \to \Delta(\mathcal{A})$ 
that maximizes the expected return:
\begin{equation}
    J(\pi) = \mathbb{E}_{\substack{g \sim p_g(\cdot), s_0 \sim \mu_0(\cdot),\\ a_t\sim\pi(\cdot| s_t, g) }}[\sum_{t=0}^{T-1} \gamma^t R(s_t, a_t, g)]
\label{eq:rl_obj}
\end{equation}


We assume the offline setting for GCRL where we aim to find the optimal policy from the dataset $\mathcal{D}$,
comprised of trajectories collected by executing a potentially sub-optimal policy. 
Each trajectory in $\mathcal{D}$ is represented by $\tau = s_{0:T}||a_{0:T}||g$,
which is the concatenation of state sequence $s_{0:T} = (s_0,\ldots,s_T)$, action sequence $a_{0:T}=(a_0,\ldots,a_T)$, and desired goal $g$, 
where $T$ is the final time-step of the trajectory.  
We assume each trajectory terminates when it achieves the goal or surpasses a sufficiently large time-step limit.
Thus, the action value $Q(s,a,g)$ represents the discounted probability of policy $\pi$ achieving the goal $g$ from the state $s$:
\begin{equation}
    \begin{split}
        Q^\pi(s,a,g) &\equiv \mathbb{E}_{\pi}[\sum_{t=0}^\infty \gamma^{t}R(s_t,a_t,g)|s_0=s,a_0=a,g] \\
        & = (1-\gamma) p^\pi(s_T = g|s_0=s, a_0=a)
    \end{split}
\end{equation}
In our approach, we aim to learn from sub-trajectory segments of length $h$+1, denoted as $s_{t:t+h} || a_{t:t+h}$, with $t \in [0, T-h]$.
We denote the reward function for sub-trajectories~\cite{chebotar2021actionable} as 
$R(s_{t:t+h},a_{t:t+h},g) \equiv \mathbf{1}(\psi(s_{t+h})=g)$, and
the space of all possible sub-trajectories as $\mathcal{SA}^h$.

\subsection{Diffusion Models for RL}
Diffusion models~\cite{ho2020denoising, song2019generative, song2020denoising, song2020score} 
are well known for effectively capturing complex distributions and generating diverse samples.
As such, they 
have recently emerged as an effective approach for GCRL~\cite{janner2022planning, li2023hierarchical}.
They have been shown to generate long-horizon plans without separate model learning, mitigating the issue of 
error accumulation often encountered in traditional planning algorithms.
Additionally, their generating process can be controlled by specifying conditions such as optimality, goal and skill.

\paragraph{Denoising Diffusion Probabilistic Model (DDPM)}
Following the prior works~\cite{janner2022planning, ajay2022conditional}, we leverage the strong capabilities of DDPM~\cite{ho2020denoising} to generate sub-trajectories of states and actions $\mathbf{x}^0 \equiv s_{t:t+h} \| a_{t:t+h} $\footnote{We use superscripts to denote the steps in the diffusion process, while subscripts to denote the time-steps in the trajectories.}.  
DDPM is composed of (1) a predefined forward noising process $q$ that incrementally adds noise to generate pure random noise
$\mathbf{x}^N \sim \mathcal{N}(\mathbf{0, I})$,
where the data is corrupted according to a pre-determined variance schedule $0 < \alpha_i < 1$, and (2)
a trainable reverse denoising process $p_\theta$ that learns to recover $\mathbf{x}^0$ from $\mathbf{x}^N$,
which yields the sampling distribution $\mathbf{x}^0\sim p_\theta(\mathbf{x}^0)$.

\begin{equation*}
\begin{split} 
    q(\mathbf{x}^{i+1}|\mathbf{x}^{i}) &\equiv \mathcal{N}(\mathbf{x}^{i+1};\sqrt{\alpha_i}\mathbf{x}^i, (1-\alpha_i)\mathbf{I}) ,
     \\
    p_\theta(\mathbf{x}^{i-1}|\mathbf{x}^i) &\equiv \mathcal{N}(\mathbf{x}^{i-1};\mu_\theta(\mathbf{x}^i, i), \mathbf{\Sigma}_i)
\end{split}
\end{equation*}
Then, the variational lower bound on log-likelihood is used as the objective function for training:
\begin{equation}  
\begin{split} \label{eq:nll}
    \mathbb{E}[-\log p_\theta(\mathbf{x}^0&)] \leq \mathbb{E}_q\big[-\log\frac{p_\theta(\mathbf{x}^{0:N})}{q(\mathbf{x}^{1:N}|\mathbf{x}^0)}\big] \\
    &= \mathbb{E}_q\big[-\log p(\mathbf{x}^N)-\sum_{i \geq 1}\log\frac{p_\theta(\mathbf{x}^{i-1}|\mathbf{x}^i)}{q(\mathbf{x}^i|\mathbf{x}^{i-1})}\big]\\
    &\equiv L(\theta) .
\end{split}
\end{equation} 
We can formulate a reparameterized version of the objective function for more efficient optimization. 
For details, we refer the readers to \cite{ho2020denoising}

\paragraph{Diffuser}
Diffuser~\cite{janner2022planning} is one of the early works on adopting diffusion models for RL, 
and showed promising results on generating high-quality plans in both offline RL and GCRL.
Diffuser is underpinned by reframing RL as a conditional sampling problem by adopting the infer-to-control 
framework. More specifically, the reverse denoising process is reformulated as an instance of classifier-guided 
sampling~\cite{sohl2015deep,dhariwal2021diffusion}, where the guide is given by the return of the sub-trajectory
\begin{equation}\label{eq:classifierguide}
    p_\theta(\mathbf{x}^{i-1}|\mathbf{x}^i,R(\mathbf{x}^0,g)) \approx \mathcal{N}(\mathbf{x}^{i-1};\mu+\Sigma \nabla R,\Sigma)
\end{equation}
where $\nabla R \equiv \nabla R(\mathbf{x}^0,g)|_{\mathbf{x}^0=\mu}$ is gradient of the return of the sub-trajectory $\mathbf{x}^0$, 
and $\mu$ and $\Sigma$ are the parameters of the plain reverse process $p_\theta(\mathbf{x}^{i-1}|\mathbf{x}^i)$.
It is worthwhile to note that when it comes to the goal-conditioned setting, Diffuser relies on inpainting to
generate trajectories that achieves the goal at the final time-step, 
rather than using the goal-dependent value function $Q(s,a,g)$
to guide the sampling process.


\subsection{Goal Relabeling Methods for GCRL}
Goal-Conditioned Reinforcement Learning (GCRL) aims to train RL agents
under a multi-goal setting~\cite{kaelbling1993learning, schaul2015universal}. 
Its offline counterpart aims to do so solely 
from the trajectory dataset without further interaction with the
environment.
One of the main challenges in offline GCRL is that, the reward signals in the dataset is extremely sparse
especially when the data collection policy is sub-optimal and rarely achieves the goal. 
Thus, in order to capture valuable skills for achieving diverse goals, goal relabeling methods are commonly employed.

\paragraph{Hindsight Experience Replay (HER)} 
Hindsight Experience Replay~\cite{andrychowicz2017hindsight} involves replacing the specified goals by 
those corresponding to states actually visited during execution. 
For example, given trajectory $\tau=s_{0:T}||a_{0:T}||g$, replacing the goal $g$ with the one corresponding to the 
final state $g'=\psi(s_T)$ yields a successful trajectory with non-zero reward.
Through such goal relabeling, we can enrich the reward signal for learning the value function, allowing us to   
extract useful skills even from unsuccessful trajectories with zero reward.
Although originally developed for the online RL setting, HER has shown to significantly improve the performance in the offline RL setting as well.

\paragraph{Actionable Model (AM)}
One shortcoming of HER is its constant generation of successful trajectories, resulting in biased training samples.
Furthermore, the relabeled goals are confined to one of the states visited within the trajectory, 
which obstructs learning to achieve a goal present in other trajectories.
More recently, AM~\cite{chebotar2021actionable} introduced a goal relabeling strategy that
generates not only successful trajectories but also unsuccessful ones.
This is done by relabeling with goal $g'$ corresponding to the state arbitrarily selected from the whole dataset.
In order to learn from unsuccessful trajectories where $g'$ is sampled from those not appearing 
in the current trajectory, AM introduces a technique called \textit{goal chaining} to examine whether $g'$ is still achievable
in some other trajectory.


This is performed by training the action-value function with the objective
\begin{equation}\label{eq:L_Q}
    L(\phi)=\mathbb{E}_{(s_t,a_t,s_{t+1})\sim \mathcal{D},g'\sim G}[(Q_\phi(s_t,a_t,g')-y(s_{t+1},g'))^2],
\end{equation}
with the goal relabeling strategy represented by $G$ and the target $y$ given by 
\begin{equation} \label{eq:AM_target}
    y(s_{t+1},g')= \begin{cases}
        1 & \text{if } \psi(s_{t})=g' \\
        \gamma\mathbb{E}_{a\sim \pi}[Q_{\phi}(s_{t+1},a,g')] & \text{otherwise}.
    \end{cases}
\end{equation}
The key idea behind the target $y$ is that even when the goal $g'$ is not immediately achievable
in the current trajectory,
there might be some other trajectory that ultimately achieves the goal, and if so, it will be reflected into the 
action-value function of state $s_{t+1}$ from that trajectory, thus effectively chaining the two trajectories 
through state $s_{t+1}$.

\section{Method}\label{sec:method}

\begin{figure}[t!]
    \centering
    \includegraphics[width=0.45\textwidth]{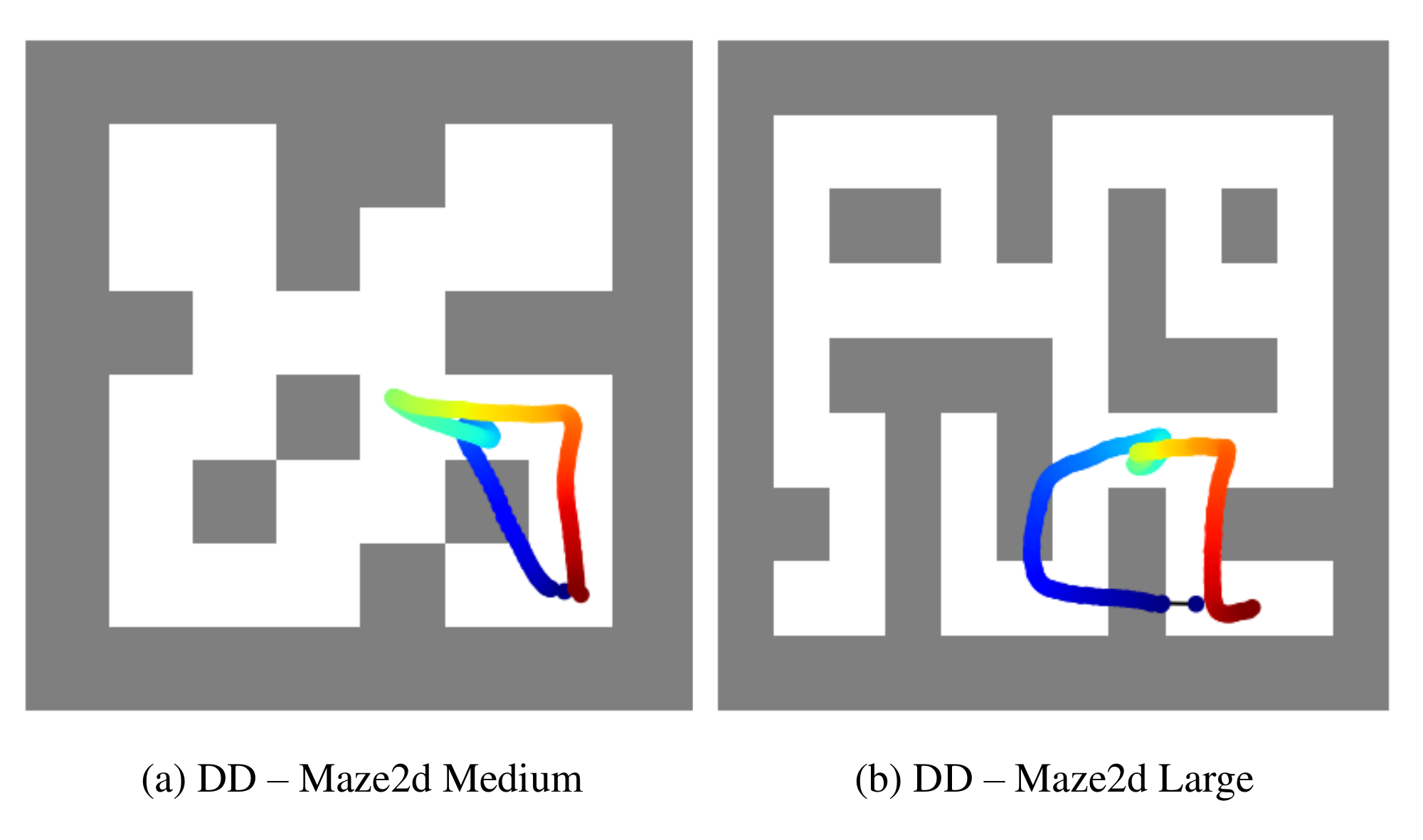}
    \caption{Plot of sampled trajectory from Decision Diffuser (DD)~\cite{ajay2022conditional}. It generates unrealistic trajectories that pass through the wall. The blue point is the initial point and the red point is the goal. left: Maze2D-Medium. right: Maze2D-Large.}
    \label{fig:maze-ablation-hit-the-wall}
\end{figure}

In this section, we introduce SSD (Sub-trajectory Stitching with Diffusion), which utilizes the conditional diffusion model to address
long-horizon tasks in offline GCRL. More concretely, SSD addresses two main challenges in offline GCRL: first, the sparsity of rewards makes 
it difficult to identify good sub-trajectories to stitch. Second, the conditional diffusion model, while holding promise in 
tackling long-horizon tasks, tends to generate unrealistic trajectories, as we show in Figure~\ref{fig:maze-ablation-hit-the-wall} for 
Decision Diffuser~(DD)~\cite{ajay2022conditional}. To overcome these challenges, SSD alternates between 
training (1) the goal-conditioned value function with multi-step goal chaining and (2) the value-conditioned diffusion model
for generating sub-plans of length $h$. 
We provide details on each part in Sections~\ref{sec:condition-prompted-unet} and \ref{sec:actionvalue}, respectively. 
The full algorithm for our approach is provided in Section~\ref{sec:alg}.

\subsection{Value-Conditioned Diffusion Model}\label{sec:condition-prompted-unet}
In order to generate more realistic sub-trajectories, we propose an architecture we name Condition-Prompted-Unet.

\begin{figure*}[t]
    \centering
    \includegraphics[width=0.95\textwidth]{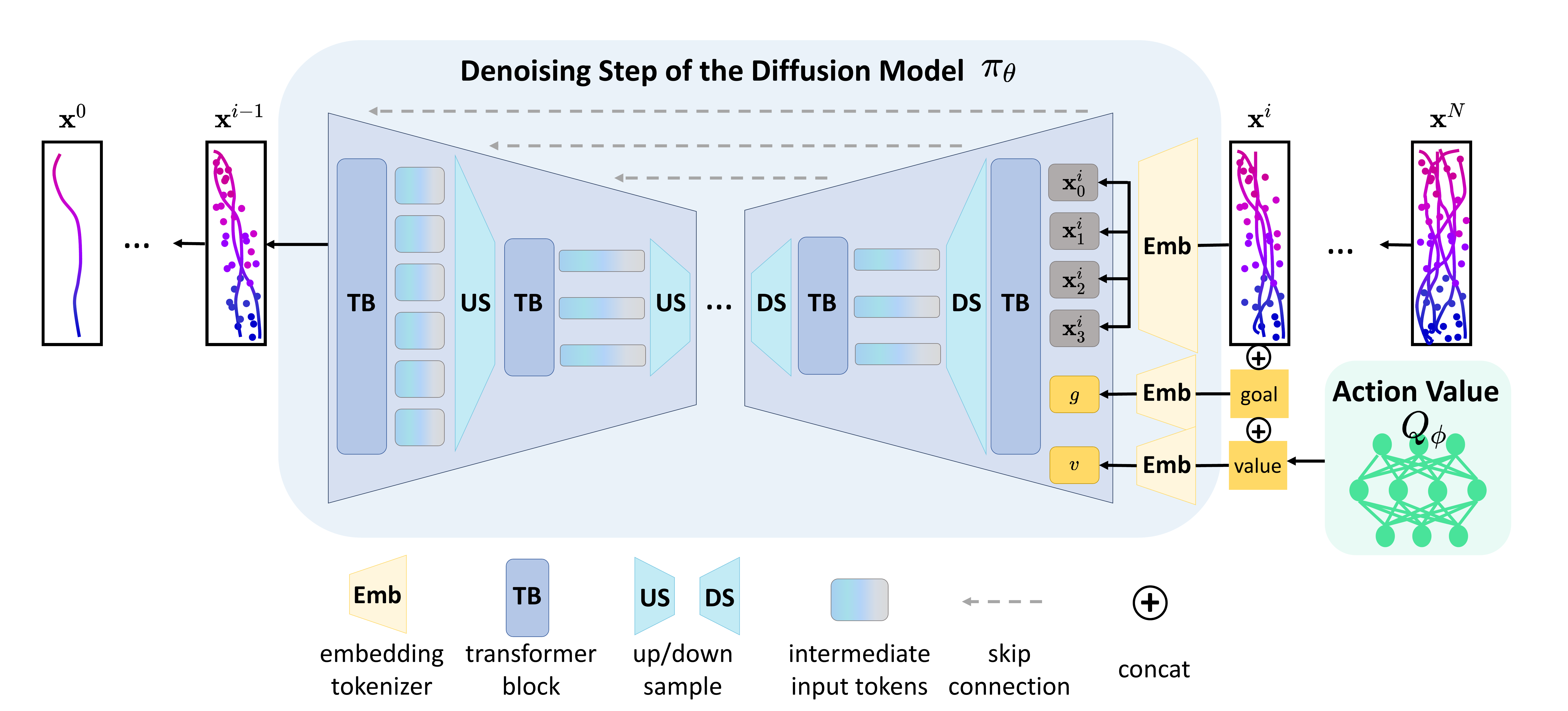}
    \caption{Architecture of the proposed Condition-Prompted-Unet model. Trajectory input is tokenized into state-action pairs. The goal and value are included as prompt tokens into the transformer layers, which constitute the Unet structure. Downsampling and Upsampling layers are integrated with each transformer block. The process is repeated for each reverse step, using the given goal and the trained action value as conditions.}
    \label{fig:architecture}
\end{figure*}

As shown in Figure~\ref{fig:architecture}, Condition-Prompted-Unet integrates the Unet architecture with transformer blocks. 
The transformer blocks provide the capacity to capture complex patterns of real-world sequential data, while the Unet structure facilitates preservation of spatial information.
By replacing convolutional blocks with transformer blocks, our approach aims to enhance the accuracy of the conditioning process.
This hybrid architecture is designed to address the challenges of realistic trajectory generation in RL domains, ensuring compatibility with environment constraints (e.g. staying within the state space and following the transition probabilities of the MDP).
The main motivation behind our model was that although the 
guidance model in Diffuser~(\eqnref{eq:classifierguide}) provides an attractive
framework to leverage return-to-go for conditional generation
of trajectory, the closed-form approximation therein was 
the main source of bottleneck for generating accurate, 
realistic trajectories. Instead, the approximation is directly learned in our model.

This conditional diffusion model $\pi_\theta$ is trained to generate 
sub-trajectories $\mathbf{x}^0 \equiv \mathbf{x}^0_{t:t+h}$
given the random noise $\mathbf{x}^N \sim \mathcal{N}(\mathbf{0}, \mathbf{I})$, the target goal $g$, and the desired probability 
specified by the target value $v$, using the reparameterized version of the
DDPM loss function in \eqnref{eq:nll}~\cite{ho2020denoising}.

\subsection{Multi-Step Goal Chaining} \label{sec:actionvalue}
Another key component of our method is the multi-step goal chaining.
Given that the diffusion model is trained with a batch of data that corresponds to sub-trajectories of length $h$+1, 
a natural goal relabeling strategy would be sampling one of the states within 
the sub-trajectory to make the sub-trajectory successful. However, sampling only within the sub-trajectory will incur a positive bias as in HER, 
thus we do so with probability 0.5, and select an arbitrary random state outside the sub-trajectory with probability 0.5. 
For the latter, we would need to perform goal chaining as in AM since it corresponds to learning from 
an unsuccessful sub-trajectory. 

A straightforward target for training the action-value function can be derived as follows: given 
sub-trajectory $\mathbf{x} \equiv s_{t:t+h}\|a_{t:t+h}$ and relabeled goal $g'$, we define target $y$ by
\begin{equation*} 
\begin{split}
    y(\mathbf{x},g') = 
     \begin{cases}
        \gamma^{k-1} \quad\quad\quad\,\, \text{if } \exists k \in [0,\ldots,h]: \psi(s_{t+k})=g' \\
        \max_{1 \le k' \le h} \gamma^{k'} \mathbb{E}_{a\sim \pi}[Q_{\phi}(s_{t+k'},a,g')] \,\,\,\, \text{o.w.}
    \end{cases}
\end{split}
\end{equation*}
This target has the following interpretation: if the relabeled goal happens to be achievable by
sampling one of the states within the sub-trajectory, we use the discounted return $\gamma^{k-1}$.
Otherwise, we attempt goal chaining at every subsequent state in the sub-trajectory and choose
the best one. 

Although this target is natural, we found that it incurs too much overestimation due to the
max operator on the bootstrapped estimates in the
offline setting. Thus, given that the sub-trajectory batch data is prepared for every time-step,
we found it sufficient to attempt goal chaining only at time-step $t$+1. In other words, we define the target $y$ by
\begin{equation} \label{eq:multistep_gc} 
\begin{split}
    y(\mathbf{x},g') = 
     \begin{cases}
        \gamma^{k-1}  \quad\quad\quad\,\, \text{if } \exists k \in [0,\ldots,h]: \psi(s_{t+k})=g' \\
        \gamma Q_{\phi}(s_{t+1},a,g') \quad\quad\quad  \text{otherwise},
    \end{cases}
\end{split}
\end{equation}
where $a \sim \pi$ is a sample from the target policy to approximate $Q_{\phi}(s_{t+1},a,g') \approx \mathbb{E}_{a\sim \pi}[Q_{\phi}(s_{t+1},a,g')]$.

\begin{algorithm*}[t]
    \caption{SSD (Training)} \label{alg:SSD}
    \begin{algorithmic}[1]
        \REQUIRE offline data $\mathcal{D}$, diffusion model $\pi_\theta$, critic $Q_\phi$, number of training iterations $K$, horizon $h$, goal relabeling strategy $G$

        \FOR{$i$ in $1,...,K$}
        \STATE $\mathbf{x}_{t:t+h} = s_{t:t+h} \| a_{t:t+h} \sim \mathcal{D}$, $g' \sim G(\mathbf{x}_{t:t+h}, \mathcal{D})$ \MYCOMMENT{sample a sub-trajectory and its relabeled goal}
        \STATE \textbf{\#\# Train action-value critic $Q_\phi$} \\
        \STATE Take a gradient descent update $\nabla_\phi L(\phi)$ using the target $y$ in \eqnref{eq:multistep_gc}, i.e.:
        \begin{align*}
            L(\phi) = & \frac{1}{2} \left[ Q_\phi(s_t, a_t, g') - y(s_{t:t+h}, a_{t:t+h}, g') \right]^2 
            , \\
            & \text{ where } y(s_{t:t+h}, a_{t:t+h}, g') = 
            \begin{cases}
            \gamma^{k-1} & \exists k \in [0,\ldots,h]: \psi(s_{t+k})=g' \\
            \gamma Q_{\phi}(s_{t+1},a \sim \pi_\theta,g') & \text{otherwise}.
            \end{cases}
        \end{align*}
        \STATE \textbf{\#\# Train diffusion model $\pi_\theta$} \\
        \IF{$\exists k \in [0,\ldots,h]: \psi(s_{t+k})=g'$} 
            \STATE $\mathbf{x}_{t:t+h} = \text{PAD}(s_{t:t+k}, h-k, s_{t+k}) \| \text{PAD}(a_{t:t+k}, h-k, a_{t+k})$ \\
             \MYCOMMENT{Goal $g'$ corresponds to state $s_{t+k}$, so replace subsequent entries by $s_{t+k}$ and $a_{t+k}$ since a successful sub-trajectory} 
        \ENDIF
        \STATE Take a gradient descent update $\nabla_\theta L(\theta)$ with 
        $\mathbf{x}_{t:t+h} \sim \pi_\theta(\boldsymbol\epsilon, g', Q_\phi(s_t,a_t,g')), \boldsymbol\epsilon \sim \mathcal{N}(\mathbf{0}, \mathbf{I})$ using the reparameterized version of \eqnref{eq:nll}
        \ENDFOR 
        \RETURN $\pi_\theta, Q_\phi$
    \end{algorithmic}
\end{algorithm*}

\subsection{Overall Algorithm} \label{sec:alg}
The pseudocode of the overall training procedure is shown in Algorithm~\ref{alg:SSD}. We use the same goal relabeling strategy $G$ 
for training the critic $Q_\phi$ and the diffusion model $\pi_\theta$, where we randomly select one of the states in the sub-trajectory with
probability 0.5, and randomly select one of the states outside the sub-trajectory with probability 0.5, as described previously. 
Although we chose uniform probability distribution for simplicity, other choices of probabilities for the goal relabeling strategy 
are certainly possible for further optimization. 

Finally, during the execution phase, we sample a trajectory $\mathbf{x}$ from $\pi_\theta$ every $k$ steps with the first state inpainted as $s_t$, and use the first $k$ sequence of $\mathbf{x}$ to take actions. This process is repeated until a goal is reached, whereby the episode terminates.

\section{Experimental Results}
\label{sec:exp}

In this section, we demonstrate the effectiveness of the proposed SSD approach in two different GCRL domains: Maze2D and Fetch. Our code is available publicly at \url{https://github.com/rlatjddbs/SSD}



\subsection{Long-Horizon Planning Tasks}
\paragraph{Maze2D} 
\begin{figure}[h]
    \centering
    \vspace{-3mm}
    \includegraphics[width=0.95\linewidth]{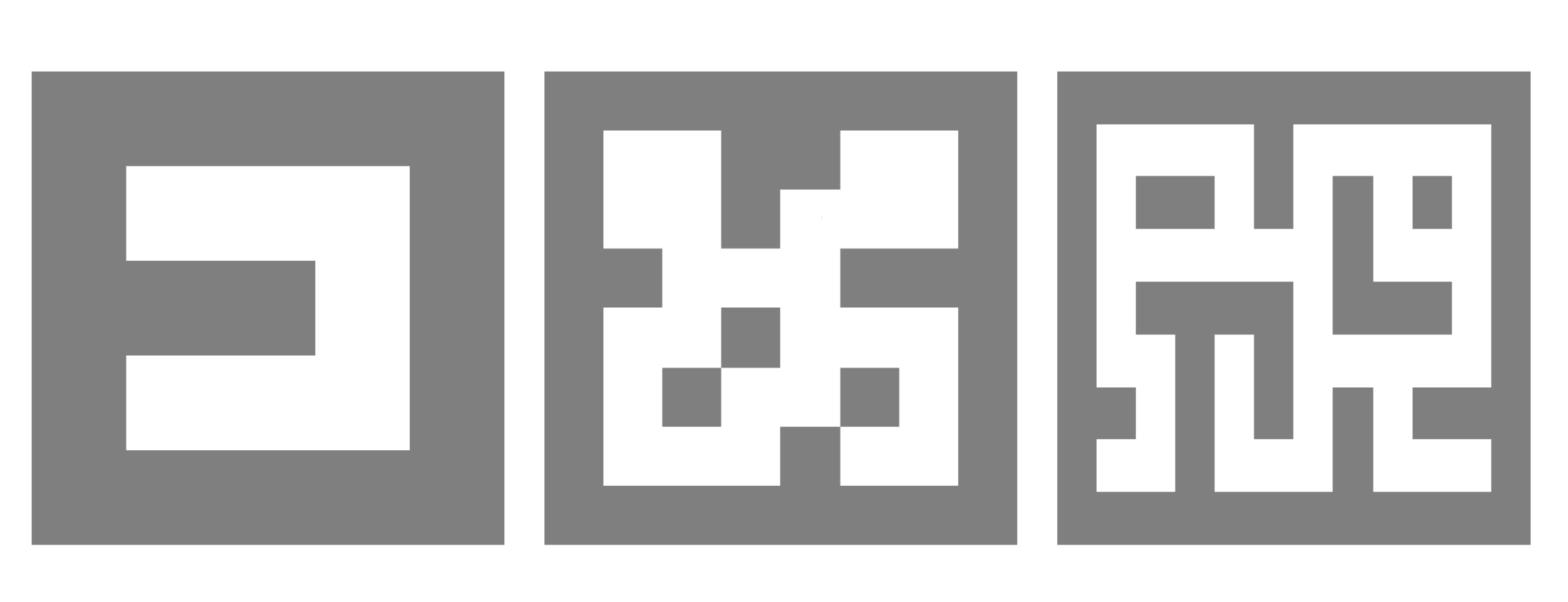}
    \caption{Layouts of Maze2d environment. left: Umaze, middle: Medium, and right: Large}
    \label{fig:maze-env}
    
\end{figure}

Maze2D is a popular benchmark for goal-oriented tasks, where a point mass agent navigates to a goal location.
These tasks are intentionally devised to assess the stitching capability of offline RL algorithms.
The environment encompasses three distinct layouts, each varying in difficulty and complexity, as depicted in Figure \ref{fig:maze-env}. Multi2D is the environment
with the same layout with multi-task setting where the goal location 
changes in every episode.

We utilize the D4RL dataset~\cite{fu2020d4rl}, which is generated by a hand-designed PID controller as a planner, which produces a sequence of waypoints. 
The dataset comprises 1 million samples for Umaze, 2 million for Medium, and 4 million for Large.

\begin{table*}[t]
\centering
\begin{tabular}{p{1.3cm}p{1.3cm}p{0.6cm}p{0.3cm}p{0.7cm}p{0.3cm}p{0.7cm}p{0.5cm}p{0.7cm}p{0.5cm}p{0.7cm}p{0.5cm}p{0.7cm}p{0.5cm}p{0.7cm}}
\toprule[1.5pt]
\multicolumn{2}{c}{\textbf{Environment}} & \multicolumn{1}{c}{\textbf{IQL}} & \multicolumn{2}{c}{\textbf{Diffuser}} & \multicolumn{2}{c}{\textbf{Diffuser}} & \multicolumn{2}{c}{\textbf{Diffuser}} & \multicolumn{2}{c}{\textbf{DD}} & \multicolumn{2}{c}{\textbf{HDMI}} &\multicolumn{2}{c}{\textbf{SSD}} \\ 
\multicolumn{2}{c}{} & \multicolumn{1}{c}{} & \multicolumn{2}{c}{} & \multicolumn{2}{c}{{\small(HER)}} & \multicolumn{2}{c}{{\small(AM)}}  & \multicolumn{2}{c}{} & \multicolumn{2}{c}{} & \multicolumn{2}{c}{{\small(Ours)}} \\ 
\midrule
Maze2D  & Umaze    & $47.4$   & $113.9$   & $\pm1.8$   & $125.4$    & $\pm0.4$  & $121.4$    & $\pm1.8$ & $116.2$ & $\pm2.7$   & $120.1$ & $\pm2.5$    & $\mathbf{144.6}$ & $\pm3.4$  \\
Maze2D  & Medium    & $34.9$   & $121.5$   & $\pm2.7$  & $130.3$    & $\pm2.3$  & $127.2$    & $\pm3.1$   & $122.3$ & $\pm2.1$   & $121.8$ & $\pm1.6$    & $\mathbf{134.4}$ & $\pm6.1$  \\
Maze2D  & Large     & $58.6$  & $123.0$   & $\pm6.4$   & $135.8$    & $\pm8.5$  & $135.0$    & $\pm3.3$ & $125.9$ & $\pm1.6$   & $128.6$ & $\pm2.9$    & $\mathbf{183.5}$ & $\pm8.6$  \\
\midrule
\multicolumn{2}{l}{Single-task Average} & $47.0$    & $119.5$  && $130.5$    && $127.9$   &&  $121.5$  &&  $123.5$  &&  $\mathbf{154.2}$  &   \\
\midrule
Multi2D & Umaze    & $24.8$   & $128.9$   & $\pm1.8$ & $132.7$    & $\pm2.4$ & $135.4$    & $\pm2.0$  & $128.2$  & $\pm2.1$   & $131.3$ & $\pm1.8$    & $\mathbf{158.2}$ & $\pm4.5$  \\
Multi2D & Medium    & $12.1$   &  $127.2$  & $\pm3.4$ & $133.0$    & $\pm2.4$  & $137.8$    & $\pm1.0$  & $129.7$  & $\pm2.7$   & $131.6$ & $\pm1.9$    & $\mathbf{155.2}$ & $\pm7.9$  \\
Multi2D & Large     & $13.9$    &  $132.1$  & $\pm5.8$ & $139.2$    & $\pm6.6$  & $145.7$    & $\pm7.7$  & $130.5$  & $\pm4.2$   & $135.4$ & $\pm2.5$    & $\mathbf{192.9}$ & $\pm8.5$  \\
\midrule
\multicolumn{2}{c}{Multi-task Average}  & $16.9$    & $129.4$   && $135.0$    && $139.6$   && $129.5$  && $132.8$  && $\mathbf{168.8}$  &   \\
\bottomrule[1.5pt]
\end{tabular}
\caption{[Maze2D] Normalized scores of Maze2D environment. Goals are fixed in Maze2D scenarios, while goals are randomly sampled in Multi2D scenarios. Reported values are mean and standard error over 5 seeds. The bolded value represents the top-performing result.}
\label{tab:maze}
\end{table*}

\paragraph{Baselines} 

Our main baseline for diffusion-based offline RL is Diffuser~\cite{janner2022planning} which is described in detail in Section~\ref{sec:prel_related}.
For a fair comparison with our value-conditioned diffusion model, 
we also extended Diffuser to condition on the action value trained with HER and AM as goal relabeling strategies (Diffuser-HER and Diffuser-AM).
Both methods use classifier guidance described in Eqn. (\ref{eq:classifierguide}).
Decision Diffuser (DD)~\cite{ajay2022conditional} is another offline RL method based on the diffusion model,
which uses classifier-free guidance and utilizes the inverse dynamics model for decision-making.
Hierarchical Diffusion for offline decision MakIng (HDMI)~\cite{li2023hierarchical} 
employs a hierarchical diffusion model involving a planner that generates sub-goals. Finally, IQL~\cite{kostrikov2021offline} is a state-of-the-art offline RL method which can be seen as advantage-weighted behavior cloning. 

\paragraph{Results}

As we show in Table~\ref{tab:maze}, SSD outperforms all baselines for all map sizes with a notable performance gap in larger maps and multi-task scenarios. 



Overall, diffusion-based algorithms all outperform IQL, which demonstrates the benefit of trajectory-wise generation of actions.

Comparing Diffuser-HER and Diffuser-AM, the AM-based guidance exhibited superior performance in multi-task scenarios where the goal switches in each episode, whereas the HER-based guidance demonstrated higher performance in simpler tasks with fixed goals. This is expected since the AM-guidance promotes stitching to 
improve handling of diverse goals.
Also, we see that employing classifier guidance in Eqn.~(\ref{eq:classifierguide}) on top of Diffuser improves the overall performance, especially when the guidance Q-value is trained based on the specified relabeling methods.
Those variants of Diffuser show even outperforming other diffusion-based algorithms such as DD and HDMI.

\begin{figure}[t!]
    \centering
    \includegraphics[width=0.5\textwidth]{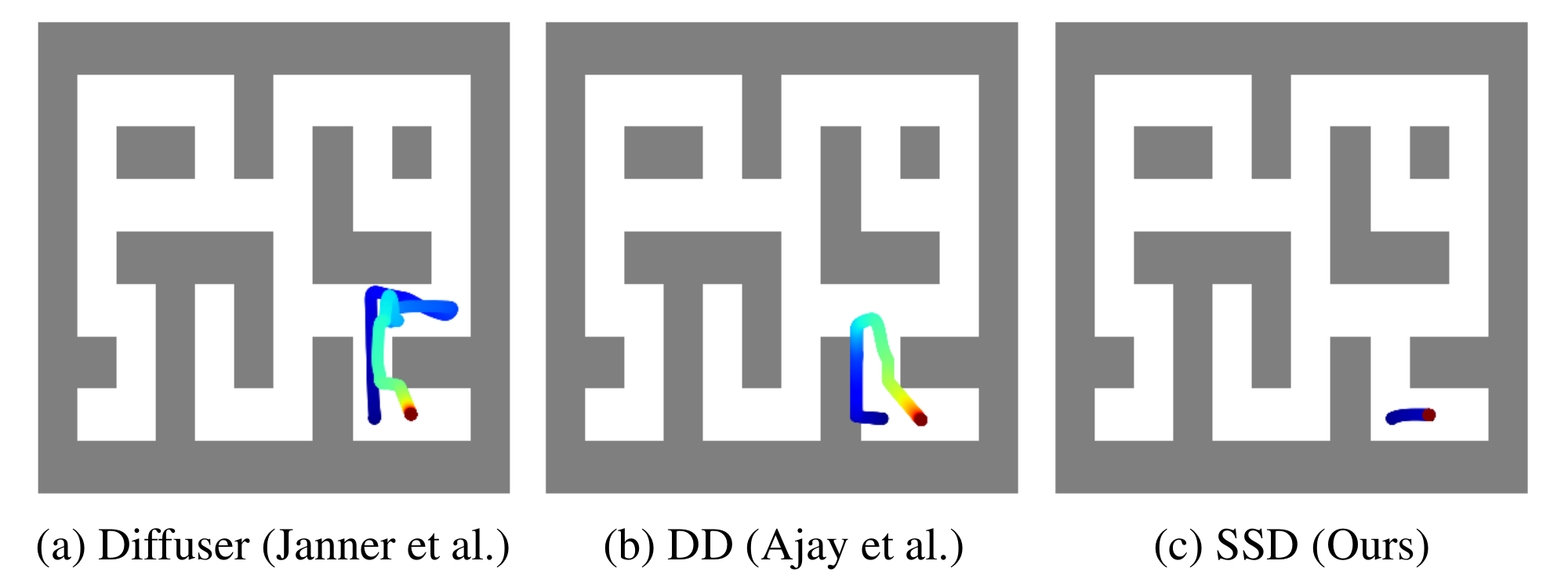}
    \caption{Comparision of behaviors of diffusion based methods. (a), (b) and (c) are result of Diffuser~\cite{janner2022planning}, DD~\cite{ajay2022conditional}, and SSD (Ours) respectively. The blue point is the initial point and the red point is the goal. All of them share the same initial point and the same goal.}
    \label{fig:maze-comparison}
\end{figure}

While all the diffusion-based policies except for HDMI aim to generate single full-trajectory in evaluation time, SSD focuses on generating sub-trajectories and stitching them. 
This property removes the necessity to know the terminal time-step $T$ a priori.
Also, we show qualitative results in Figure~\ref{fig:maze-comparison} demonstrating that the baseline methods can produce unnecessary detouring trajectories instead of taking a more direct path as SSD does. 



\subsection{Robotic Manipulation Tasks}
\begin{table*}[t]
\centering
\begin{tabular}{p{3.0cm}|p{0.45cm}p{0.75cm}p{0.45cm}p{0.75cm}p{0.45cm}p{0.75cm}|p{0.45cm}p{0.75cm}p{0.45cm}p{0.75cm}p{0.45cm}p{0.75cm}p{0.45cm}p{0.75cm}}
\toprule[1.5pt]
\multicolumn{1}{c|}{\textbf{Environment}} & \multicolumn{6}{c|}{\textbf{Supervised Learning}} & \multicolumn{6}{c}{\textbf{Q-value Based}} \\ 
\multicolumn{1}{c|}{} & \multicolumn{2}{c}{GCSL} & \multicolumn{2}{c}{{WGCSL}} & \multicolumn{2}{c|}{{GoFAR}} & \multicolumn{2}{c}{AM} & \multicolumn{2}{c}{DDPG} & \multicolumn{2}{c}{SSD(Ours)} \\ 
\midrule
FetchReach  & $0.98$ & {\scriptsize$\pm0.05$}    & $0.99$  & {\scriptsize$\pm0.01$}    & $\mathbf{1.0}$    & {\scriptsize$\pm0.0$}   & $\mathbf{1.0}$  & {\scriptsize$\pm0.0$}  & $0.99$    & {\scriptsize$\pm0.02$}  & $\mathbf{1.0}$  & {\scriptsize$\pm0.0$}  \\
FetchPickAndPlace   & $0.54$ & {\scriptsize$\pm0.20$}    & $0.54$  & {\scriptsize$\pm0.16$}    & $0.84$    & {\scriptsize$\pm0.12$}   & $0.78$   & {\scriptsize$\pm0.15$}  & $0.81$   & {\scriptsize$\pm0.13$}    & $\mathbf{0.9}$ & {\scriptsize$\pm0.03$}  \\
FetchPush   & $0.72$ & {\scriptsize$\pm0.15$}    & $0.76$  & {\scriptsize$\pm0.12$}    & $0.88$    & {\scriptsize$\pm0.09$}   & $0.67$   & {\scriptsize$\pm0.14$}  & $0.65$   & {\scriptsize$\pm0.18$}   &  $\mathbf{0.97}$ & {\scriptsize$\pm0.02$}   \\
FetchSlide  & $0.17$ & {\scriptsize$\pm0.13$}    & $0.18$  & {\scriptsize$\pm0.14$}    & $0.18$    & {\scriptsize$\pm0.12$}   & $0.11$  & {\scriptsize$\pm0.09$}  & $0.08$    & {\scriptsize$\pm0.11$}  &   $0.1$ & {\scriptsize$\pm0.07$}  \\
\midrule 
Average     & $0.60$    & & $0.62$   & & $0.73$  & & $0.64$  & & $0.63$  & & $\mathbf{0.74}$  & \\
\bottomrule[1.5pt]
\end{tabular}
\caption{[Fetch] Success rate in Fetch environment. Reported values are mean and standard error over 5 seeds. The bolded value represents the top-performing result.}
\label{tab:fetch_sr}
\end{table*}

\begin{table*}[h]
\centering
\begin{tabular}{p{3.0cm}|p{0.45cm}p{0.75cm}p{0.45cm}p{0.75cm}p{0.45cm}p{0.75cm}|p{0.45cm}p{0.75cm}p{0.45cm}p{0.75cm}p{0.45cm}p{0.75cm}p{0.45cm}p{0.75cm}}
\toprule[1.5pt]
\multicolumn{1}{c|}{\textbf{Environment}} & \multicolumn{6}{c|}{\textbf{Supervised Learning}} & \multicolumn{6}{c}{\textbf{Q value Based}} \\ 
\multicolumn{1}{c|}{} & \multicolumn{2}{c}{GCSL} & \multicolumn{2}{c}{{WGCSL}} & \multicolumn{2}{c|}{{GoFAR}} & \multicolumn{2}{c}{AM} & \multicolumn{2}{c}{DDPG} & \multicolumn{2}{c}{SSD(Ours)} \\ 
\midrule
FetchReach  & $20.91$   & {\scriptsize$\pm2.78$}    & $21.9$  & {\scriptsize$\pm2.13$}    & $28.2$    & {\scriptsize$\pm0.61$}    & $\mathbf{30.1}$  & {\scriptsize$\pm0.32$}  & $29.8$    & {\scriptsize$\pm0.59$}  & $29.18$  & {\scriptsize$\pm0.32$}  \\
FetchPickAndPlace   & $8.94$    & {\scriptsize$\pm3.09$}    & $9.84$  & {\scriptsize$\pm2.58$}    & $\mathbf{19.7}$    & {\scriptsize$\pm2.57$}    & $18.4$    & {\scriptsize$\pm3.51$}  & $16.8$    & {\scriptsize$\pm3.10$}  & $18.0$   & {\scriptsize$\pm0.25$}  \\
FetchPush   & $13.4$    & {\scriptsize$\pm3.02$}    & $14.7$  & {\scriptsize$\pm2.65$}    & $18.2$    & {\scriptsize$\pm3.00$}    & $14.0$    & {\scriptsize$\pm2.81$}  & $12.5$    & {\scriptsize$\pm4.93$}  & $\mathbf{19.84}$  & {\scriptsize$\pm0.50$}  \\
FetchSlide  & $1.75$    & {\scriptsize$\pm1.3$}     & $\mathbf{2.73}$  & {\scriptsize$\pm1.64$}    & $2.47$    & {\scriptsize$\pm1.44$}    & $1.46$    & {\scriptsize$\pm1.38$}  & $1.08$    & {\scriptsize$\pm1.35$}  &  $2.6$ & {\scriptsize $\pm 0.94$}  \\  
\midrule
Average     & $11.25$ & & $12.29$   & & $17.14$ & & $16.00$ & & $15.05$ & & $\mathbf{17.40}$ &  \\
\bottomrule[1.5pt]
\end{tabular}
\caption{[Fetch] Discounted return in Fetch environment. Reported values are mean and standard error over 5 different training seeds. The bolded value represents the top-performing result. }
\label{tab:fetch_dr}
\end{table*}

\paragraph{Baselines} 
GCSL~\cite{ghosh2019learning} combines behavior cloning with hindsight goal relabeling to make trajectories optimal. 
WGCSL~\cite{yang2022rethinking} additionally uses a weighting scheme, by considering the importance of relabeled goals. 
GoFAR~\cite{ma2022far} is a \textit{relabeling-free} method, 
formulating GCRL as a state-occupancy matching problem. 
DDPG refers to the DDPG~\cite{lillicrap2019continuous} with hindsight-relabeling~\cite{andrychowicz2017hindsight} to
learn a goal-conditioned critic.

\paragraph{Results}
We compare our approach to the aforementioned baseline methods in the Fetch environment, 
a robotic manipulation environment which includes FetchReach, FetchPush, FetchPickAndPlace, and FetchSlide. 
We use the offline dataset provided in \citet{ma2022far}.

As our proposed method is formulated primarily to achieve the goal as soon as possible, we employ two quantitative metrics, namely success rate and discounted return, to rigorously evaluate and quantify its performance.
As shown in Table \ref{tab:fetch_sr}, SSD achieves the state-of-the-art performance in average success rate across all tasks, 
by a large margin especially in FetchPush and FetchPickAndPlace. 
Also, SSD exhibits the state-of-the-art performance in average discounted return across all tasks, as shown in Table \ref{tab:fetch_dr}.

\subsection{Performance Impact of Target Value $v$}
\begin{figure}[h!]
    \centering
    \includegraphics[width=0.4\textwidth]{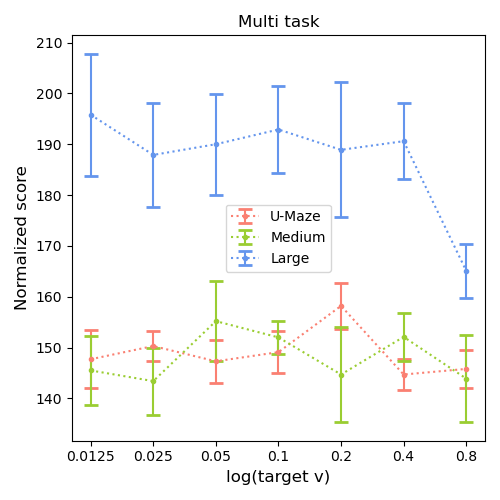}
    \caption{Plot of normalized scores across varying target values. Umaze domain achieves the highest score at $v=0.2$, Medium domain at $v=0.05$, and Large domain at $v=0.0125$}
    \label{fig:maze-ablation-score}
\end{figure}

Here, we provide additional results which show the performance across different values of $v$.
When deployed to the environment, our algorithm chooses actions based on the target value $v$,
akin to the return-to-go in Decision Transformer~\cite{chen2021decision}.
The target value $v$ represents the discounted probability of achieving a goal $g$ from the current state.
Since we want to achieve the goal as quickly as possible, we should assign a higher value to $v$ at test time.
However, an appropriate value for $v$ depends on the complexity of the task.
As shown in Figure \ref{fig:maze-ablation-score}, 
the target value for achieving the best performance decreases as the task complexity increases from Umaze to Large.
This result implies that the target value conditioning is generally affecting the plan generation as desired.

\section{Conclusion}
\label{sec:conclusion}
In this work, we have presented a novel approach named SSD to address the challenges encountered in Offline GCRL, especially in long-horizon tasks. 
First, our proposed approach offers effective solutions to overcome the issue of sparse rewards by generating sub-trajectories that are conditioned on action values. 
These action values are obtained through our novel hindsight relabeling technique, which enhances the stitching of trajectories. 
Second, our approach integrates the training of action values and conditioning within the diffusion training process, eliminating the need for separate guidance training. This integration minimizes approximation errors and leads to improved performance, as demonstrated in the Maze2D and Multi2D results.
The experimental results validate the effectiveness of SSD and highlight its potential for practical applications in the field of GCRL.

\section*{Acknowledgments}
This work was supported by the IITP grant funded by MSIT (No.2020-0-00940, Foundations
of Safe Reinforcement Learning and Its Applications to Natural Language Processing; No.2022-0-
00311, Development of Goal-Oriented Reinforcement Learning Techniques for Contact-Rich Robotic
Manipulation of Everyday Objects; No.2019-0-00075, AI Graduate School Program (KAIST)).



\bibliography{main-new}
\clearpage
\appendix
\section{Related Works} 
\label{appx:related}
\subsection{Offline GCRL Methods}
\subsubsection{Goal-Conditioned Supervised Learning (GCSL).} 

Goal-Conditioned Supervised Learning (GCSL)~\cite{ghosh2019learning} is a straightforward approach involving supervised imitation learning with hindsight-relabeled datasets. 
They leverage the stability of supervised imitation learning, eliminating the requirement for expert demonstrations. 
They consider the hindsight relabeled dataset as an \textit{expert} demonstration. 
Also, they theoretically analyze the objective function being optimized by GCSL, and show it is equivalent to optimizing a lower bound on the RL objective. 

\subsubsection{Weighted GCSL.}

Subsequently, \citet{yang2022rethinking} introduced an improved method called Weighted GCSL (WGCSL), specifically tailored for offline GCRL.
WGCSL introduces a weighting scheme that considers the significance of samples or trajectories to learn optimal policy.
They present the objective function of WGCSL as follows:
\begin{equation}\label{eq:WGCSL}
    J_{WGCSL}(\pi)= \mathbb{E}_{\substack{g\sim p_g(\cdot),\tau \sim P_{\mathcal{D}}(\cdot), \\ t\sim [1,T], k\sim[t,T]}}\big[w_{t,k} \log \pi_\theta (a_t|s_t,\psi(s_k)) \big]    
\end{equation}
where $w_{t,k}=\gamma^{k-t}$. 
This weight is named \textit{discounted relabeling weight}. 
Equation \ref{eq:WGCSL} implies that WGCSL resembles supervised learning, while any $(k-t)$ step later states after $s_t$ can be utilized as relabeled goals.
Then they show that the surrogate of RL objective is lower bounded by the objective of the WGCSL, with a tighter bound compared to GCSL.
\begin{equation}
    \begin{split}
        J_{surr}(\pi) \geq T\cdot J_{WGCSL}(\pi) \geq T\cdot J_{GCSL}(\pi),
    \end{split}
\end{equation}
where
\begin{equation}
    \begin{split}
    J_{surr}(\pi)=\frac{1}{T}\mathbb{E}_{\substack{g\sim p_g(\cdot), \\ \tau \sim P_\mathcal{D}(\cdot)}} \big[&\sum_{t=0}^T \log \pi(a_t|s_t,g) \cdot \\
    &\sum_{k=t}^T \gamma^{k-1} R(s_t,a_t,g) \big]    
    \end{split}
\end{equation}
which is the surrogate of Equation \ref{eq:rl_obj}.

Moreover, they improve the weighting scheme by introducing two additional weights. 
Firstly, \textit{goal-conditioned exponential advantage weight} is defined as $h(s,a,g) = \exp(A(s,a,g) + C)$, which assigns greater weight to more significant state-action-goal samples evaluated by the universal value function.
The constant $C$ ensures that $h(s,a,g) \geq 1$.

Secondly, \textit{best-advantage weight} is defined as:
\begin{equation}
    \epsilon(A(s_t,a_t,\psi(s_k)))=\begin{cases}
        1, &A(s_t,a_t,\psi(s_k)) > \hat{A} \\
        \epsilon_{min} &\text{otherwise}
    \end{cases}
\end{equation}
Here, $\hat{A}$ is a threshold that gradually increases and $\epsilon_{min}$ is a small positive value. 
This mechanism progressively guides the policy towards a global optimum based on the learned value function. 
Finally, they combine these weights in the following form: $w_{t,k} = \gamma^{k-t} \exp(A(s_t,a_t,\psi(s_k)) + C) \cdot \epsilon(A(s_t,a_t,\psi(s_k)))$.

\subsubsection{Goal-Conditioned f-Advantage Regression (GoFAR).}
GoFAR~\cite{ma2022far} presents a new paradigm of \textit{relabeling-free} method optimizing the same goal-conditioned RL objective Equation \ref{eq:rl_obj}.
They formulate GCRL problem as a state occupancy matching problem. 
Goal-conditioned state-action occupancy distribution $d^\pi (s,a;g):\mathcal{S}\times\mathcal{A}\times\mathcal{G}\to[0,1]$ of $\pi$ is
\begin{equation}
    d^\pi(s,a;g)\equiv (1-\gamma)\sum_{t=0}^\infty \gamma^t Pr(s_t=s,a_t=a|s_0, a_t \sim \pi(s_t,g), s_{t+1})
\end{equation}
Then the state-occupancy distribution is $d^\pi(s;g)=\sum_a d^\pi(s,a;g)$. Finally, the policy is as follows: $\pi(a|s,g)=\frac{d^\pi(s,a;g)}{d^\pi(s;g)}$.
Also, joint goal-state-action density induced by $p_g(g)$ and the policy $\pi$ is $d^\pi(s,g)=p_g(g)d^\pi(s,a;g)$.
Finally, when $d^\pi$ is given, we can rewrite Equation \ref{eq:rl_obj} as below:
\begin{equation}
    J(\pi)=\frac{1}{1-\gamma}\mathbb{E}_{(s,g)\sim d^\pi(s,g),a\sim\pi(a|s)}[R(s,a,g)]
\end{equation}

In this context, they introduce an offline lower bound through the incorporation of an $f$-divergence regularization term. The formulation is as follows:
\begin{center}
\begin{equation}\label{eq:far_KL}
    \begin{split}
        &-D_{KL}(d^\pi(s;g)||p(s;g)) \geq \\
        &\mathbb{E}_{(s,g)\sim d^\pi(s,g)}[\log p(s;g)] - D_f (d^\pi (s,a;g)||d^O(s,a;g))
    \end{split}
\end{equation}
\end{center}
Furthermore, they demonstrate that the dual problem of Eq. (\ref{eq:far_KL}) can be transformed into an unconstrained optimization problem. 
This yields the optimal goal-weighting distribution, which consequently allows GoFAR to treat all offline data as though it originates from the optimal policy. 
This attribute is obtained solely by addressing the state occupancy matching problem, obviating the necessity for explicit hindsight relabeling.

\subsection{Reinforcement Learning with Diffusion Models}
\subsubsection{Decision Diffuser (DD).}
Distinguishing itself from Diffuser~\cite{janner2022planning}, \citet{ajay2022conditional} adopted a different approach by employing \textit{classifier-free guidance}~\cite{ho2022classifier} instead of using classifier guidance, and incorporating inverse dynamics.
While Diffuser involves the training of an additional classifier, \citet{ajay2022conditional} highlighted that this introduces complexities in the training process. 
Moreover, potential approximation errors can arise from both the use of pre-trained classifiers, which may be trained on noisy data, and the approximation of the guided-reverse process as a Gaussian process (as depicted in Equation (\ref{eq:classifierguide})).
By omitting the supplementary classifier guidance, which can be accompanied by approximation errors in the value function, the approach referred to as DD demonstrates superior performance in MuJoCo locomotion tasks like hopper, halfcheetah, and walker2d.
However, this method has not been extended to the realm of GCRL, which encompasses tasks with sparse rewards and therefore necessitates effective stitching capabilities.
Consequently, it still lacks the proficiency to seamlessly combine or stitch distinct segments of trajectories, a requirement crucial for tasks in the GCRL domain.

\subsubsection{DiffusionQL.}
Meanwhile, \citet{wang2022diffusion} formulate diffusion models as a policy that generates action.
They straightforwardly interpret the diffusion objective as a behavior cloning objective, while adding a negative Q value as a regularization component to the diffusion loss.
\begin{equation}
    L(\theta)=L_{ddpm}(\theta)-\alpha\mathbb{E}_{s\sim P_{\mathcal{D}}, a^0\sim \pi_\theta}[Q_\phi (s,a^0)]
\end{equation}
where $a^0$ is the output of the reverse process. 
The Q-function $Q_\phi$ and the Diffusion policy $\pi_\theta$ undergo iterative training, with the Q-learning process following the methodology outlined in \citet{kumar2020conservative}. 
It's important to note that the gradient backpropagation of Q-learning with respect to the action goes through the entire diffusion process.
Although the results demonstrate impressive performance in MuJoCo locomotion tasks, the approach lacks an extension to the long-horizon setting of GCRL problems.

\subsubsection{AdaptDiffuser.}
\citet{liang2023adaptdiffuser} introduce an extended task-oriented diffusion model for planning purposes, which is extended based on \cite{janner2022planning}'s approach. 
They highlight a critical issue associated with the classifier guidance mechanism represented by Equation \ref{eq:classifierguide}. 
They point out that if $\mu$ in Equation \ref{eq:classifierguide} significantly deviates from the optimal trajectory, regardless of how strong the guidance $g$ is, the result will be highly biased.
This bias can be particularly pronounced when dealing with imbalanced datasets across different tasks, potentially resulting in poor adaptation performance on unseen tasks. 
To address this challenge, they propose a solution involving the generation of reward-guided synthetic data. 
This data augmentation strategy, integrated with the diffusion model, enhances AdaptDiffuser's ability to adapt to more intricate tasks beyond those encountered during training.

\subsubsection{Hierarchical Diffusion for Offline Decision Making (HDMI).}
Additionally, \citet{li2023hierarchical} address the challenges of long-horizon offline RL by implementing a hierarchical architecture. 
First, since they leverage hierarchical architecture, they conduct subgoal extraction inspired by the method presented in SoRB~\cite{eysenbach2019search}.
This approach utilizes graphic abstractions of the environment to automatically identify and define subgoals.
Secondly, they propose a novel architecture for the diffusion model, moving away from the conventional Unet-based design.
Instead, they adopt a transformer-based structure, based on diffusion transformer blocks~\cite{peebles2023scalable}.
Leveraging the dataset enriched with subgoal constraints and the transformer-based diffusion architecture, they proceed to train the diffusion model using a classifier-free guidance approach. 

\begin{center}
    \begin{table*}[h]
        \centering
        \begin{tabular}{p{1.3cm}p{1.3cm}p{0.5cm}p{0.7cm}p{0.5cm}p{0.7cm}p{0.5cm}p{0.7cm}p{0.5cm}p{0.7cm}}
\toprule[1.5pt]
\multicolumn{2}{c}{\textbf{Environment}} & \multicolumn{2}{c}{Diffuser(Org)} & \multicolumn{2}{c}{Diffuser(Re)} & \multicolumn{2}{c}{Diffuser(HER)} & \multicolumn{2}{c}{Diffuser(AM)}\\ 
\midrule
Maze2d  & U-Maze    & $113.9$  & {\scriptsize$\pm1.8$} & $111.3$  & {\scriptsize$\pm1.5$}  & $125.4$  & {\scriptsize$\pm0.4$}    & $121.4$  & {\scriptsize$\pm1.8$}         \\
Maze2d  & Medium    & $121.5$  & {\scriptsize$\pm2.7$} & $122.2$  & {\scriptsize$\pm2.0$}    & $130.3$  & {\scriptsize$\pm2.3$}    & $127.2$  & {\scriptsize$\pm3.1$}   \\
Maze2d  & Large     & $123.0$  & {\scriptsize$\pm6.4$}   & $137.0$  & {\scriptsize$\pm5.1$}   & $135.8$  & {\scriptsize$\pm8.5$}  & $135.0$  & {\scriptsize$\pm3.3$}     \\
\midrule
Multi2d & U-Maze    & $128.9$  & {\scriptsize$\pm1.8$}   & $128.8$  & {\scriptsize$\pm2.1$}   & $132.7$  & {\scriptsize$\pm2.4$}   & $135.4$  & {\scriptsize$\pm2.0$}   \\
Multi2d & Medium    & $127.2$  & {\scriptsize$\pm3.4$}     & $133.4$  & {\scriptsize$\pm1.8$}     & $133.0$  & {\scriptsize$\pm2.4$}    & $137.8$  & {\scriptsize$\pm1.0$}   \\
Multi2d & Large     & $132.1$  & {\scriptsize$\pm5.8$}   & $143.2$  & {\scriptsize$\pm12.2$}   & $139.2$  & {\scriptsize$\pm6.6$} & $145.7$  & {\scriptsize$\pm7.7$}      \\
\bottomrule[1.5pt]
        \end{tabular}
        \caption{Reproduced results of Diffuser, Diffuser(HER) and Diffuser(AM)}
        \label{tab:ablation_diffuser}
    \end{table*}
\end{center}

\section{Reproduction of Diffuser}
We reproduced Diffuser~\cite{janner2022planning} for implementation of Diffuser-HER and Diffuser-AM in Table~\ref{tab:maze}. The first column, labeled Diffuser(Org), displays the original values reported by \citet{janner2022planning}. The second column, Diffuser(Re), presents the reproduced values obtained through our own implementation. The subsequent columns showcase the performance of Diffuser-HER and Diffuser-AM, respectively.

\section{Details of Experimental Environments and Offline Datasets} \label{appx:env}
\subsection{Maze2d}
\begin{figure}[ht]
    \centering
    \includegraphics[width=0.5\textwidth]{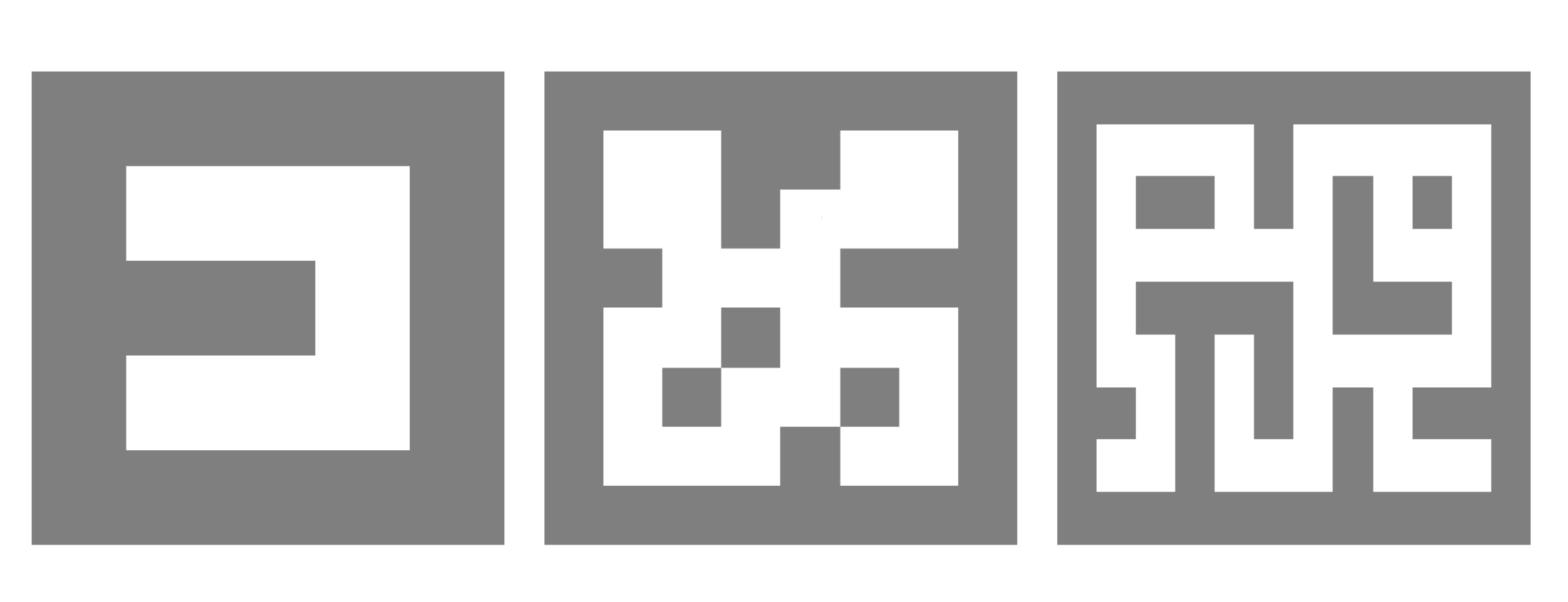}
    \caption{Layouts of Maze2d environment. \textbf{left:} Maze2d-Umaze, \textbf{middle:} Maze2d-Medium, and \textbf{right:} Maze2d-Large}
    \label{fig:maze-env}
\end{figure}

Maze2d environment is a representative goal-oriented task, where a point mass agent is navigating to find a fixed goal location.
These tasks are intentionally devised to assess the stitching capability of offline RL algorithms.
The environment encompasses three distinct layouts, each varying in difficulty and complexity, as depicted in Figure \ref{fig:maze-env}.

We employ D4RL dataset~\cite{fu2020d4rl}, an offline collection of trajectory samples. 
This dataset is generated using a hand-designed PID controller as a planner, which produces sequences of waypoints. 
The data collection process unfolds as follows: commencing from an initial position, waypoints are strategically planned using goal information and the data collection policy. 
When the goal is attained, a new goal is sampled, and this cycle persists.
The dataset comprises 1 million samples for Umaze, 2 million for Medium, and 4 million for Large.

\subsection{Fetch}
\begin{figure}[ht]
    \centering
    \includegraphics[width=0.48\textwidth]{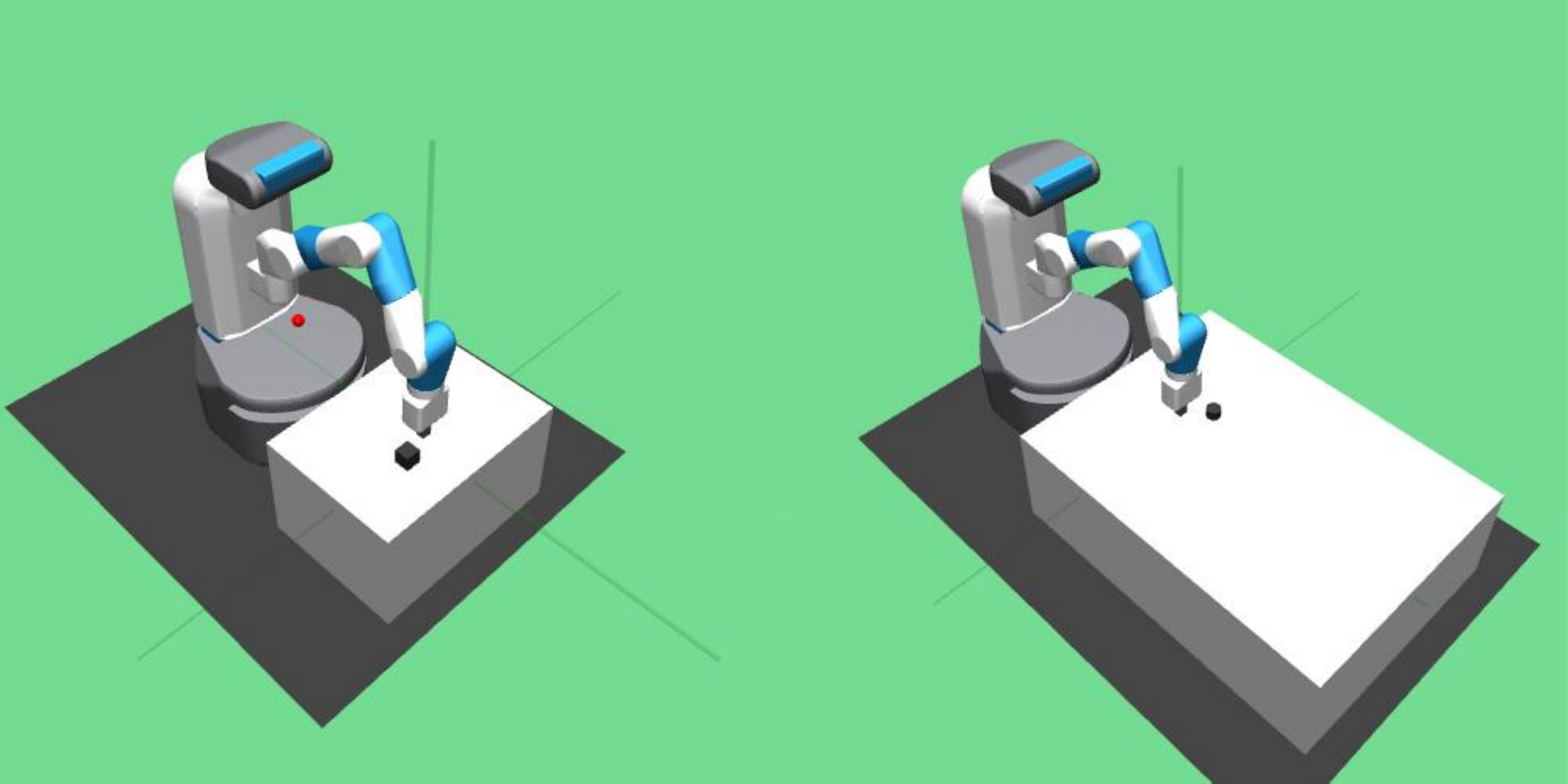}
    \caption{Visualizations of Fetch environment domain. \textbf{left:} FetchPush and FetchPickAndPlace, \textbf{right:} FetchSlide}
    \label{fig:fetch-env}
\end{figure}

The Fetch Robotics environment encompasses robot manipulation tasks available in the OpenAI Gym. 
These manipulation tasks are notably more challenging than the continuous control environments in MuJoCo. 
The environment features tasks like Reach, Push, PickAndPlace, and Slide, which closely simulate real-world scenarios. 

The Fetch environments are characterized by the following specifications:
\begin{itemize}
    \item Robotic arm with seven degrees of freedom
    \item Robotic arm with a two-pronged parallel gripper
    \item Cartesian coordinates for 3D goal representation
    \item Sparse reward signal: 1 then at goal with tolerance 5cm and 1 otherwise.
    \item Four-dimensional action space: three Cartesian dimensions and one dimension to control the gripper
    \item MuJoCo physics engine to provide information about joint angles, velocities, object positions, rotations, and velocities
\end{itemize}

In the Reach task, the Fetch robot arm is tasked with relocating its end-effector to a specified goal location. 
In the Push task, the robot arm is required to move a box by pushing it toward a designated goal position. 
The PickAndPlace task involves the robot arm grasping a box from a table using its gripper and subsequently transferring it to a predetermined goal location, either on the table or suspended in the air.
In the Slide task, the robot arm is engaged in striking a puck across a longer table, with the objective of causing the puck to slide and ultimately reach a specified goal location.

For offline training, we adopt the offline dataset generation approach proposed by \citet{ma2022far}. 
They provide 'random' and 'expert' datasets each of which is generated by a random policy agent and a trained DDPG-HER agent. 
Following their approach, we create a blended dataset by combining 900K transitions from the random dataset and 100K transitions from the expert dataset except for FetchReach. 
For FetchReach, we used the 'random' dataset exclusively.
This hybrid setup has also been adopted in previous studies~\cite{kim2022demodice, ma2022versatile}, and it has been demonstrated to be realistic as only a small fraction of the dataset is task-specific, while all transitions provide valuable insights into the environment.

\section{Implementation Details}
\subsection{Action Selection Procedure}
During evaluation, it is important how to utilize generated sub-trajectory to take actions. We choose suitable technique to take action for each environment.  
Similar to conventional planning-based RL methods, we execute actions by utilizing the initial part of the planned sub-trajectory.
\subsubsection{Maze2d}
In the Maze2d experiment, we use inverse dynamics, which follows the technique from \citet{janner2022planning, ajay2022conditional, li2023hierarchical}. 
First we estimate the inverse dynamics of the environment $F(s_t, s_{t+1})$ which is inference of action $a_t$ given current state $s_t$ and next state $s_{t+1}$.
Leveraging only state information of the first half of each generated sub-trajectory, we can obtain actions with $F$ from the state sequences.
As a result, we generate a trajectory for every $\frac{h}{2}$ step. 
Furthermore, once the agent reaches the goal, it takes actions to minimize the distance between the target goal and the current state, as described in the approach by \citet{janner2022planning}.

\subsubsection{Fetch}
In the Fetch experiment, we simply utilize only the first action of each generated trajectory.
As a result, we generate a trajectory for every step, in a manner similar to conventional planning-based RL methods~\cite{allgower2012nonlinear, janner2022planning, ajay2022conditional}. 
Once the agent reaches the goal, the generation process stops and the agent maintains its current state.

\subsection{Main Hyperparameters}
\begin{table}[h]
    \centering
    \begin{tabular}{cccc}
        \toprule[1.5pt]
         \multicolumn{2}{c}{Environment} & \multicolumn{1}{c}{Horizon} &\multicolumn{1}{c}{Target $v$} \\
         \midrule
         Maze2d & U-maze    & 64    & 0.2   \\
         Maze2d & Medium    & 64    & 0.05  \\
         Maze2d & Large     & 64    & 0.1  \\
         \midrule
         Fetch  & Reach      & 16    & 1.0  \\
         Fetch  & Pick\&Place& 16    & 0.2  \\
         Fetch  & Push       & 16    & 0.025  \\
         Fetch  & Slide      & 16    & 0.05  \\
         \bottomrule[1.5pt]
         
    \end{tabular}
    \caption{Main hyperparameters: horizon(training time) and target value(evaluation time).}
    \label{tab:hyperparam}
\end{table}

\subsection{Other Hyperparameters}
As described in Table~\ref{tab:other_hyperparam}, for maze2d experiments, we use 1M of training iteration $K$, 50 of diffusion steps $N$, 1.0 of action weight, $0.98$ of discount factor $\gamma$, $2e-4$ of learning rate $\lambda$. For fetch experiments, we use 2M of training iteration $K$, 50 of diffusion steps $N$, 10.0 of action weight, $0.98$ of discount factor $\gamma$, $2e-4$ of learning rate $\lambda$.

\begin{table}[H]
    \centering
    \begin{tabular}{c|ccccc}
        \toprule[1.5pt]
         \multicolumn{1}{c|}{Environment}    & \multicolumn{1}{c}{K}    & \multicolumn{1}{c}{N} &\multicolumn{1}{c}{action weight}    &\multicolumn{1}{c}{$\gamma$}  &\multicolumn{1}{c}{$\lambda$}   \\
         \midrule
         Maze2d    & $1M$     & 50    & 1.0    & 0.98   & $2e-4$   \\
         \midrule
         Fetch     & $2M$     & 50    & 10.0   & 0.98   & $2e-4$  \\
         \bottomrule[1.5pt]
         
    \end{tabular}
    \caption{The number of iteration $K$, diffusion steps $N$, action weight, discounted factor $\gamma$ and $\lambda$}
    \label{tab:other_hyperparam}
\end{table}

\subsection{Hyperparameters for Diffuser-HER and Diffuer-AM (Table~\ref{tab:maze})}

We used the same hyperparameter with the official code\footnote{https://github.com/jannerm/diffuser}: 
\begin{itemize}
    \item training step: $2M$
    \item guide scale: $0.1$
    \item guide steps: $2$
\end{itemize}

\begin{table*}[h!]
\centering
\begin{tabular}{p{2.0cm}|p{0.45cm}p{0.75cm}p{0.45cm}p{0.75cm}p{0.45cm}p{0.75cm}|p{0.45cm}p{0.75cm}p{0.45cm}p{0.75cm}p{0.45cm}p{0.75cm}p{0.45cm}p{0.75cm}}
\toprule[1.5pt]
\multicolumn{1}{c|}{\textbf{Environment}} & \multicolumn{6}{c|}{\textbf{Supervised Learning}} & \multicolumn{6}{c}{\textbf{Q value Based}} \\ 
\multicolumn{1}{c|}{} & \multicolumn{2}{c}{GCSL} & \multicolumn{2}{c}{{WGCSL}} & \multicolumn{2}{c|}{{GoFAR}} & \multicolumn{2}{c}{AM} & \multicolumn{2}{c}{DDPG} & \multicolumn{2}{c}{SSD(Ours)} \\ 
\midrule
FetchReach  & $0.008$ & {\scriptsize$\pm0.008$}    & $\mathbf{0.007}$  & {\scriptsize$\pm0.004$}    & $0.018$    & {\scriptsize$\pm0.003$}   & $\mathbf{0.007}$  & {\scriptsize$\pm0.001$}   & $0.041$    & {\scriptsize$\pm0.005$}  & $0.029$  & {\scriptsize$\pm0.001$}  \\
FetchPick   & $0.108$ & {\scriptsize$\pm0.060$}    & $0.094$  & {\scriptsize$\pm0.043$}    & $\mathbf{0.036}$    & {\scriptsize$\pm0.013$}   & $0.040$   & {\scriptsize$\pm0.020$}  & $0.043$    & {\scriptsize$\pm0.021$}    & $0.062$ & {\scriptsize$\pm0.010$}  \\
FetchPush   & $0.042$ & {\scriptsize$\pm0.018$}    & $0.041$  & {\scriptsize$\pm0.020$}    & $0.033$    & {\scriptsize$\pm0.008$}   & $0.070$   & {\scriptsize$\pm0.039$}  & $0.060$    & {\scriptsize$\pm0.026$}   & $\mathbf{0.029}$ & {\scriptsize$\pm0.001$}   \\
FetchSlide  & $0.204$ & {\scriptsize$\pm0.05$}     & $0.173$  & {\scriptsize$\pm0.04$}     & $\mathbf{0.120}$    & {\scriptsize$\pm0.02$}    & $0.198$   & {\scriptsize$\pm0.06$}   & $0.353$    & {\scriptsize$\pm0.25$}  & $0.375$& {\scriptsize$\pm0.019$}  \\
\midrule
Average    & $0.091$    & & $0.079$ & & $\mathbf{0.052}$ & & $0.079$ & & $0.124$   & & $0.124$ &  \\
\bottomrule[1.5pt]
\end{tabular}
\vspace{5pt}
\caption{[Fetch] Final distance. Reported values are mean and standard error over 5 different training seeds. The \textbf{bolded} value represents the top-performing result. }
\label{tab:fetch_fd_appx}
\end{table*}

\begin{table*}[h!]
\centering
\begin{tabular}{p{1.3cm}p{1.3cm}p{0.5cm}p{0.7cm}p{0.5cm}p{0.7cm}p{0.5cm}p{0.7cm}p{0.5cm}p{0.7cm}p{0.5cm}p{0.7cm}p{0.5cm}p{0.7cm}p{0.5cm}p{0.7cm}}
\toprule[1.5pt]
\multicolumn{2}{c}{\textbf{Environment}} & \multicolumn{14}{c}{Target value} \\ 
\multicolumn{2}{c}{} & \multicolumn{2}{c}{$v=0.0125$} & \multicolumn{2}{c}{$v=0.025$} & \multicolumn{2}{c}{{$v=0.05$}} & \multicolumn{2}{c}{{$v=0.1$}} & \multicolumn{2}{c}{$v=0.2$} & \multicolumn{2}{c}{$v=0.4$} & \multicolumn{2}{c}{$v=0.8$} \\ 
\midrule
Maze2d  & U-Maze    & $\mathbf{142.1}$  & {\scriptsize$\pm5.8$}    & $\mathbf{134.7}$  & {\scriptsize$\pm6.9$}    & $\mathbf{142.1}$  & {\scriptsize$\pm3.4$}    & $\mathbf{149.1}$    & {\scriptsize$\pm5.4$}   & $\mathbf{144.6}$  & {\scriptsize$\pm3.4$}   & $\mathbf{150.3}$    & {\scriptsize$\pm7.4$}   & $\mathbf{135.7}$    & {\scriptsize$\pm6.9$}     \\
Maze2d  & Medium    & $\mathbf{131.4}$  & {\scriptsize$\pm5.4$}    & $\mathbf{137.3}$ & {\scriptsize$\pm5.5$}    & $\mathbf{134.4}$  & {\scriptsize$\pm6.1$}    & $\mathbf{134.5}$    & {\scriptsize$\pm3.2$}   & $\mathbf{129.5}$   & {\scriptsize$\pm5.0$}  & $\mathbf{128.2}$    & {\scriptsize$\pm4.6$}   & $\mathbf{130.1}$    & {\scriptsize$\pm6.8$}    \\
Maze2d  & Large     & $\mathbf{171.4}$  & {\scriptsize$\pm7.7$}     & $\mathbf{170.6}$ & {\scriptsize$\pm4.2$}   & $\mathbf{184.9}$  & {\scriptsize$\pm9.8$}   & $\mathbf{183.5}$    & {\scriptsize$\pm8.6$}  & $\mathbf{175.2}$   & {\scriptsize$\pm12.7$}  & $\mathbf{161.3}$    & {\scriptsize$\pm8.8$}   & $\mathbf{160.4}$ & {\scriptsize$\pm12.1$}    \\
\midrule
Multi2d & U-Maze    & $\mathbf{147.7}$  & {\scriptsize$\pm5.7$}    & $\mathbf{150.3}$ & {\scriptsize$\pm3.0$}   & $\mathbf{147.3}$  & {\scriptsize$\pm4.2$}    & $\mathbf{149.1}$   & {\scriptsize$\pm4.1$}   & $\mathbf{158.2}$  & {\scriptsize$\pm4.5$}   & $\mathbf{144.7}$    & {\scriptsize$\pm3.0$}   & $\mathbf{145.8}$    & {\scriptsize$\pm3.8$}    \\
Multi2d & Medium    & $\mathbf{145.5}$  & {\scriptsize$\pm6.7$}    & $\mathbf{143.4}$ & {\scriptsize$\pm6.6$}    & $\mathbf{155.2}$  & {\scriptsize$\pm7.9$}    & $\mathbf{152.0}$    & {\scriptsize$\pm3.3$}   & $\mathbf{144.7}$   & {\scriptsize$\pm9.4$}  & $\mathbf{152.1}$    & {\scriptsize$\pm4.7$}   & $\mathbf{143.9}$    & {\scriptsize$\pm8.6$}    \\
Multi2d & Large     & $\mathbf{195.8}$  & {\scriptsize$\pm12.0$}     & $\mathbf{187.9}$ & {\scriptsize$\pm10.2$}    & $\mathbf{190.0}$  & {\scriptsize$\pm9.9$}     & $\mathbf{192.9}$    & {\scriptsize$\pm8.5$}   & $\mathbf{188.9}$   & {\scriptsize$\pm13.3$}  & $\mathbf{190.6}$    & {\scriptsize$\pm7.5$}   & $\mathbf{165.1}$ & {\scriptsize$\pm5.3$}   \\
\bottomrule[1.5pt]
\end{tabular}
\vspace{5pt}
\caption{[Maze2d] Normalized scores across target values. Reported values are mean and standard error over 5 different training seeds. The \textbf{bolded} value represents the state-of-the-art result. We choose a target value $v$ for each environment that optimizes the average performance across both single and multi-task scenarios.}
\label{tab:maze_ablation}
\end{table*}
\section{Omitted Experimental Results}

As shown in Table \ref{tab:fetch_sr}, SSD achieves the state-of-the-art performance in average success rate across all tasks, 
by a large margin especially in FetchPush and FetchPickAndPlace. 


\subsection{Fetch Missing Results} \label{appx:fetch}
Our approach aims to arrive at the goal as fast as possible, however, has no capability to reduce the distance once achieved the goal. 
As a result, the success rate and discounted return show a state-of-the-art level on average (Table~\ref{tab:fetch_dr} and Table~\ref{tab:fetch_sr}), while the performance falls behind in terms of final distance (Table~\ref{tab:fetch_fd_appx}).
Even in the most challenging task, Slide, SSD exhibits a competitive level of discounted return.

\subsection{Quantitative and Qualitative Analysis} 
\label{appx:ablation}

\paragraph{Quantitative Analysis}

\begin{figure}[h!]
    \centering
    \includegraphics[width=\linewidth]{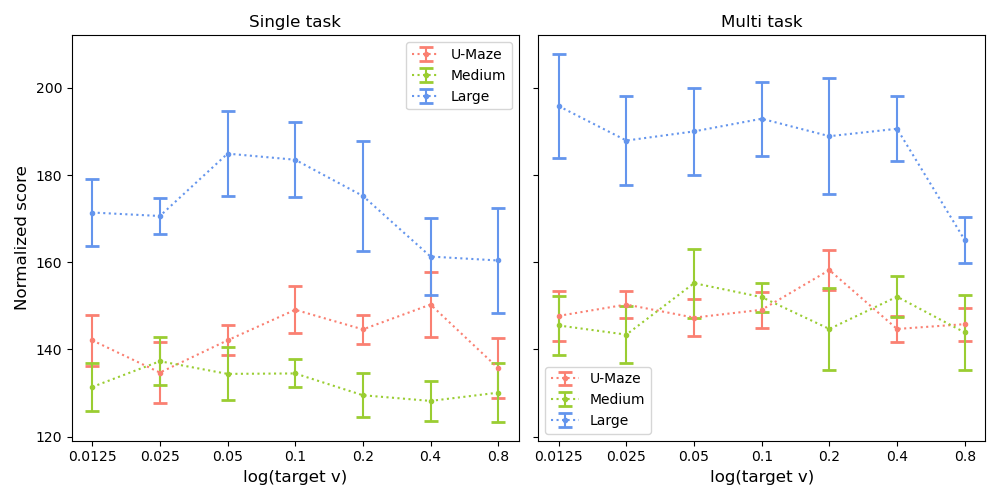}
    \caption{[Maze2d] Plot of normalized score and standard error with a logarithmic scale on the x-axis representing the target value. \textbf{left:} single-task scenario and \textbf{right:} multi-task scenario.}
    \label{fig:maze-ablation-graph}
\end{figure}

In the Maze2d experiment, SSD demonstrates its state-of-the-art performance across all target values (Table \ref{tab:maze_ablation}). 
However, distinct peak points can be observed for each task, indicating varying levels of difficulty (Figure \ref{fig:maze-ablation-graph}).
In the single-task scenario, Umaze, Medium, and Large achieve their highest scores at $v=0.4$, $v=0.025$, and $v=0.05$, respectively. 
In the multi-task scenario, the peak points shift to $v=0.2$, $v=0.05$, and $v=0.0125$ for Umaze, Medium, and Large, respectively.
It is evident that the optimal target value decreases as the maze size increases, indicating that more challenging tasks necessitate lower target values.

In the Fetch experiment, a similar trend based on the target value can be observed. 
Each task demonstrates variations in its optimal target value, reflecting its level of difficulty.
The success rate, final distance, and discounted return metrics also support these trends (Table \ref{tab:fetch_ablation_sr}, Table \ref{tab:fetch_ablation_fd}, and Table \ref{tab:fetch_ablation_dr}).
With respect to the success rate, FetchReach, FetchPickAndPlace, FetchPush and FetchSlide show their peak point at $v=1$, $v=0.2$, $v=0.025$, and $v=0.05$, each of which represents its difficulty level as well. 

\begin{table*}[t]
\centering
\begin{tabular}{p{2.8cm}|p{0.45cm}p{0.7cm}p{0.45cm}p{0.7cm}p{0.45cm}p{0.7cm}p{0.45cm}p{0.7cm}p{0.45cm}p{0.7cm}p{0.45cm}p{0.7cm}p{0.45cm}p{0.7cm}}
\toprule[1.5pt]
\multicolumn{1}{c|}{\textbf{Environment}} & \multicolumn{14}{c}{Target value} \\ 
\multicolumn{1}{c|}{} & \multicolumn{2}{c}{$v=0.025$} & \multicolumn{2}{c}{$v=0.05$} & \multicolumn{2}{c}{$v=0.1$} & \multicolumn{2}{c}{$v=0.2$} & \multicolumn{2}{c}{$v=0.4$} & \multicolumn{2}{c}{$v=0.8$} & \multicolumn{2}{c}{$v=1.0$} \\ 
\midrule
FetchReach  & $0.6$   & {\scriptsize$\pm0.03$}    & $0.8$  & {\scriptsize$\pm0.01$}    & $0.4$    & {\scriptsize$\pm0.05$}    & $0.8$  & {\scriptsize$\pm0.03$}  & $0.6$    & {\scriptsize$\pm0.02$}  & $\mathbf{1}$  & {\scriptsize$\pm0.0$} & $\mathbf{1}$  & {\scriptsize$\pm0.0$}  \\
FetchPickAndPlace   & $0.58$    & {\scriptsize$\pm0.05$}    & $0.64$  & {\scriptsize$\pm0.09$}    & $0.8$    & {\scriptsize$\pm0.06$}    & $\mathbf{0.9}$    & {\scriptsize$\pm0.03$}  & $0.81$    & {\scriptsize$\pm0.02$}  & $0.83$   & {\scriptsize$\pm0.05$}    & $0.84$   & {\scriptsize$\pm0.04$}  \\
FetchPush   & $\mathbf{0.97}$    & {\scriptsize$\pm0.02$}    & $\mathbf{0.93}$  & {\scriptsize$\pm0.03$}    & $\mathbf{0.94}$    & {\scriptsize$\pm0.02$}    & $\mathbf{0.9}$    & {\scriptsize$\pm0.05$}  & $\mathbf{0.96}$    & {\scriptsize$\pm0.02$}  & $0.75$  & {\scriptsize$\pm0.07$}   & $0.74$  & {\scriptsize$\pm0.07$}  \\
FetchSlide  & $0.07$    & {\scriptsize$\pm0.02$}     & $0.1$  & {\scriptsize$\pm0.04$}    & $0.07$    & {\scriptsize$\pm0.03$}    & $0.07$    & {\scriptsize$\pm0.04$}  & $0.05$    & {\scriptsize$\pm0.04$}  &  $0.0$ & {\scriptsize $\pm0.0$} &  $0.0$ & {\scriptsize $\pm0.0$}  \\  
\bottomrule[1.5pt]
\end{tabular}
\vspace{5pt}
\caption{[Fetch] Final success rate across target values. Reported values are mean and standard error over 5 different training seeds. The \textbf{bolded} value indicates that its error bound overlaps with the state-of-the-art result.}
\label{tab:fetch_ablation_sr}
\end{table*}

\begin{table*}[h!]
\centering
\begin{tabular}{p{2.8cm}|p{0.45cm}p{0.7cm}p{0.45cm}p{0.7cm}p{0.45cm}p{0.7cm}p{0.45cm}p{0.7cm}p{0.45cm}p{0.7cm}p{0.45cm}p{0.7cm}p{0.45cm}p{0.7cm}}
\toprule[1.5pt]
\multicolumn{1}{c|}{\textbf{Environment}} & \multicolumn{14}{c}{Target value} \\ 
\multicolumn{1}{c|}{} & \multicolumn{2}{c}{$v=0.025$} & \multicolumn{2}{c}{$v=0.05$} & \multicolumn{2}{c}{$v=0.1$} & \multicolumn{2}{c}{$v=0.2$} & \multicolumn{2}{c}{$v=0.4$} & \multicolumn{2}{c}{$v=0.8$} & \multicolumn{2}{c}{$v=1.0$} \\ 
\midrule
FetchReach  & $0.077$   & {\scriptsize$\pm0.001$}    & $0.063$  & {\scriptsize$\pm0.004$}    & $0.067$    & {\scriptsize$\pm0.004$}    & $0.052$  & {\scriptsize$\pm0.003$}  & $0.090$    & {\scriptsize$\pm0.001$}  & $0.032$  & {\scriptsize$\pm0.002$} & $0.027$  & {\scriptsize$\pm0.001$}  \\
FetchPickAndPlace   & $0.149$    & {\scriptsize$\pm0.011$}    & $0.132$  & {\scriptsize$\pm0.031$}    & $0.084$    & {\scriptsize$\pm0.025$}    & $0.062$    & {\scriptsize$\pm0.01$}  & $0.116$    & {\scriptsize$\pm0.024$}  & $0.095$   & {\scriptsize$\pm0.021$}    & $0.096$   & {\scriptsize$\pm0.016$}  \\
FetchPush   & $\mathbf{0.029}$    & {\scriptsize$\pm0.001$}    & $0.041$  & {\scriptsize$\pm0.008$}    & $0.037$    & {\scriptsize$\pm0.007$}    & $0.060$    & {\scriptsize$\pm0.011$}  & $0.043$    & {\scriptsize$\pm0.01$}  & $0.149$  & {\scriptsize$\pm0.04$}   & $0.117$  & {\scriptsize$\pm0.031$}  \\
FetchSlide  & $0.476$    & {\scriptsize$\pm0.056$}     & $0.353$  & {\scriptsize$\pm0.05$}    & $0.479$    & {\scriptsize$\pm0.05$}    & $0.414$    & {\scriptsize$\pm0.04$}  & $0.48$    & {\scriptsize$\pm0.03$}  &  $0.519$ & {\scriptsize $\pm0.03$}    &  $0.567$ & {\scriptsize $\pm0.04$}  \\  
\bottomrule[1.5pt]
\end{tabular}
\vspace{5pt}
\caption{[Fetch] Final distance across target values. Reported values are mean and standard error over 5 different training seeds. The \textbf{bolded} value indicates that its error bound overlaps with the state-of-the-art result.}
\label{tab:fetch_ablation_fd}
\end{table*}

\begin{table*}[h!]
\centering
\begin{tabular}{p{2.8cm}|p{0.45cm}p{0.7cm}p{0.45cm}p{0.7cm}p{0.45cm}p{0.7cm}p{0.45cm}p{0.7cm}p{0.45cm}p{0.7cm}p{0.45cm}p{0.7cm}p{0.45cm}p{0.7cm}}
\toprule[1.5pt]
\multicolumn{1}{c|}{\textbf{Environment}} & \multicolumn{14}{c}{Target value} \\ 
\multicolumn{1}{c|}{} & \multicolumn{2}{c}{$v=0.025$} & \multicolumn{2}{c}{$v=0.05$} & \multicolumn{2}{c}{$v=0.1$} & \multicolumn{2}{c}{$v=0.2$} & \multicolumn{2}{c}{$v=0.4$} & \multicolumn{2}{c}{$v=0.8$} & \multicolumn{2}{c}{$v=1.0$} \\ 
\midrule
FetchReach  & $14.24$   & {\scriptsize$\pm0.46$}    & $19.17$  & {\scriptsize$\pm0.36$}    & $11.44$    & {\scriptsize$\pm0.92$}    & $18.36$  & {\scriptsize$\pm0.89$}  & $15.36$    & {\scriptsize$\pm0.79$}  & $26.14$  & {\scriptsize$\pm0.92$} & $\mathbf{29.82}$  & {\scriptsize$\pm0.84$}  \\
FetchPick   & $9.62$    & {\scriptsize$\pm1.02$}    & $11.50$  & {\scriptsize$\pm1.43$}    & $14.25$    & {\scriptsize$\pm0.77$}    & $17.95$    & {\scriptsize$\pm0.25$}  & $15.59$    & {\scriptsize$\pm1.10$}  & $18.04$   & {\scriptsize$\pm1.15$}    & $17.95$   & {\scriptsize$\pm1.24$}  \\
FetchPush   & $17.49$    & {\scriptsize$\pm0.95$}    & $15.64$  & {\scriptsize$\pm0.83$}    & $\mathbf{18.94}$    & {\scriptsize$\pm0.91$}    & $17.37$    & {\scriptsize$\pm1.14$}  & $\mathbf{19.84}$    & {\scriptsize$\pm0.50$}  & $16.2$  & {\scriptsize$\pm1.33$}   & $16.08$  & {\scriptsize$\pm1.87$}  \\
FetchSlide  & $1.66$    & {\scriptsize$\pm0.57$}     & $2.04$  & {\scriptsize$\pm0.49$}    & $1.25$    & {\scriptsize$\pm0.34$}    & $\mathbf{2.47}$    & {\scriptsize$\pm1.01$}  & $1.24$    & {\scriptsize$\pm0.64$}  &  $0.32$ & {\scriptsize $\pm0.16$} &  $0.57$ & {\scriptsize $\pm0.11$}  \\  
\bottomrule[1.5pt]
\end{tabular}
\vspace{5pt}
\caption{[Fetch] Discounted return across target values. Reported values are mean and standard error over 5 different training seeds. The \textbf{bolded} value indicates that its error bound overlaps with the state-of-the-art result.}
\label{tab:fetch_ablation_dr}
\end{table*}

However, the FetchSlide task presents a unique challenge for SSD due to its requirement to maintain the object's location relative to its velocity. 
SSD's design, which assumes the episode's completion upon goal achievement, results in a relatively high rate of successful trajectories based on the discounted return metric (Table \ref{tab:fetch_ablation_dr}). 
While the final success rate (Table \ref{tab:fetch_ablation_sr}) might be low, the competitive discounted return indicates that SSD successfully reaches the goal but might overshoot it.

\begin{figure*}[ht]
    \centering
    \includegraphics[width=\textwidth]{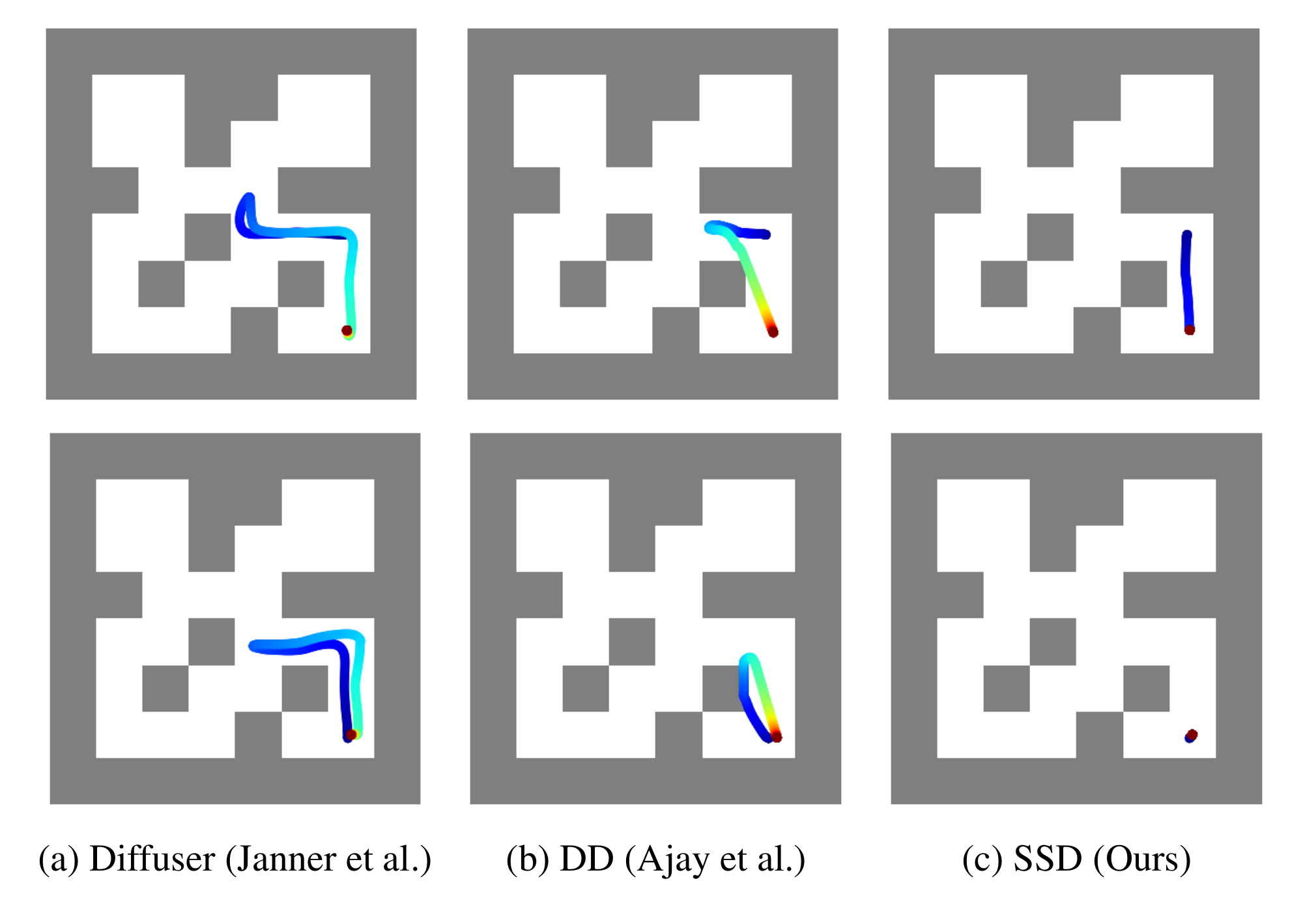}
    \caption{Maze2d-Medium. The blue point is the initial point and the red point is the goal. Each row shares the same initial point. (a), (b) and (c) are result of Diffuser~\cite{janner2022planning}, DD~\cite{ajay2022conditional}, and SSD(Ours) respectively. }
    \label{fig:maze-ablation-medium}
\end{figure*}

\begin{figure*}[ht]
    \centering
    \includegraphics[width=\textwidth]{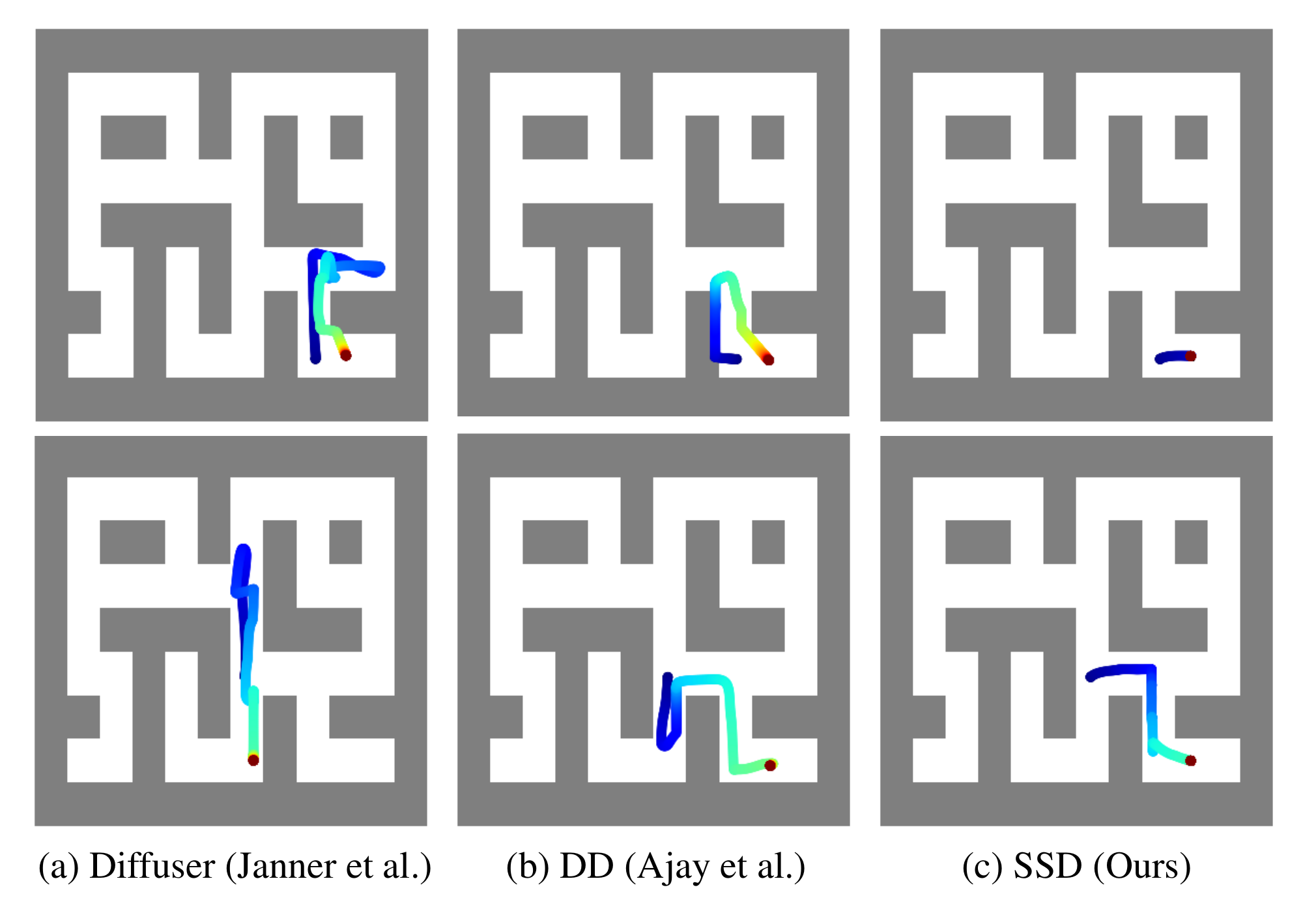}
    \caption{Maze2d-Large. The blue point is the initial point and the red point is the goal. Each row shares the same initial point. (a), (b) and (c) are result of Diffuser~\cite{janner2022planning}, DD~\cite{ajay2022conditional}, and SSD(Ours) respectively.}
    \label{fig:maze-ablation-large}
\end{figure*}

\begin{figure*}[ht]
    \centering
    \includegraphics[width=0.5\textwidth]{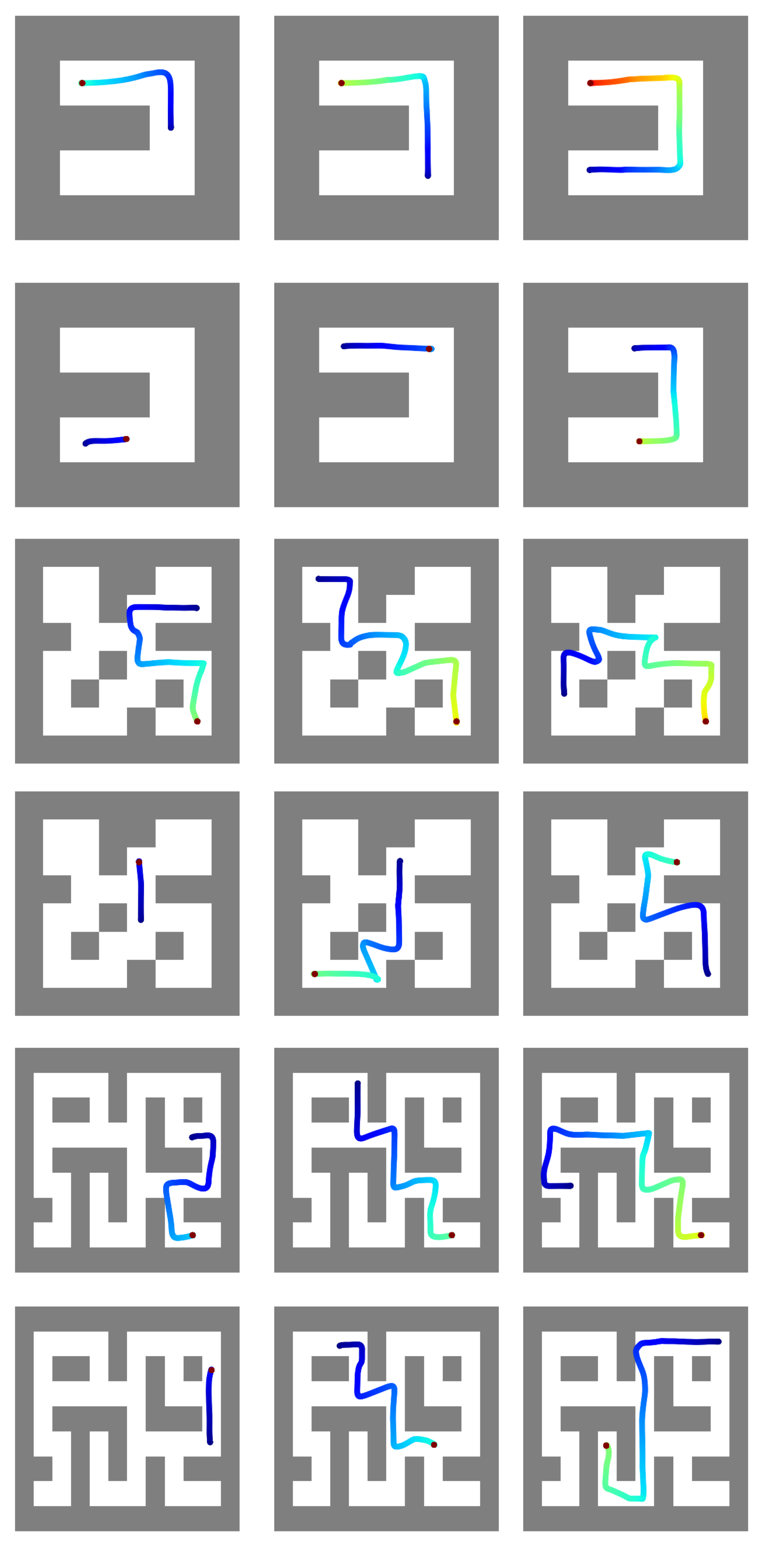}
    \caption{Random visualizations of the evaluation results. The rows correspond to different scenarios: Umaze (single-task), Umaze (multi-task), Medium (single-task), Medium (multi-task), Large (single-task), and Large (multi-task).}
    \label{fig:maze-generation}
\end{figure*}

\begin{figure*}[ht]
    \centering
    \includegraphics[width=\textwidth]{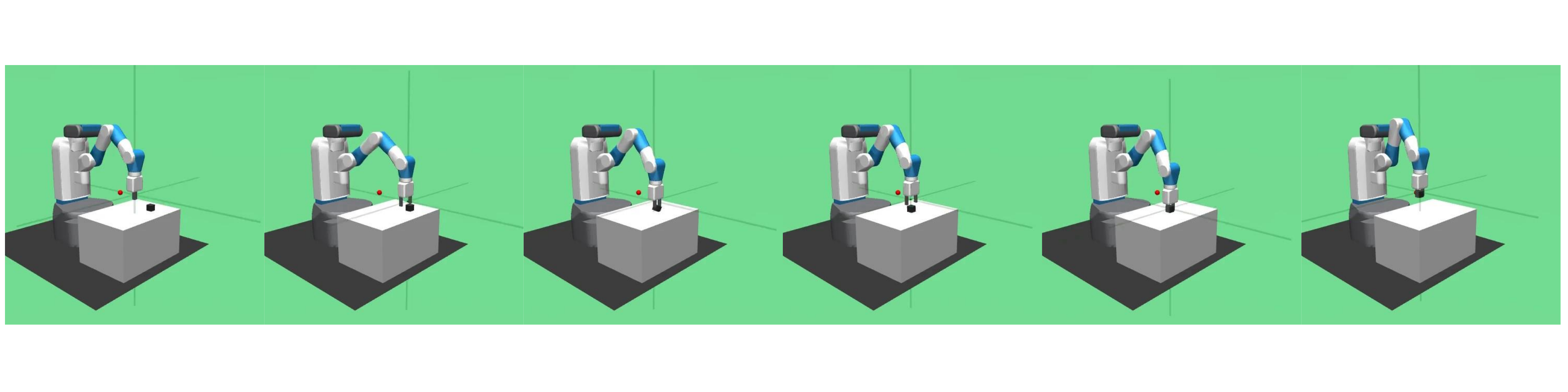}
    \caption{Random visualizations of the evaluation result. FetchPickAndPlace.}
    \label{fig:fetch-generation}
\end{figure*}

\paragraph{Qualitative Analysis}
In contrast to other diffusion-based planning methods, SSD follows a more direct path to the goal, while the others tend to loop away from it (Figure \ref{fig:maze-ablation-medium}, Figure \ref{fig:maze-ablation-large}).
SSD consistently performs well by identifying the fastest route regardless of the starting point or the goal location (Figure \ref{fig:maze-generation}).
We show random visualization of FetchPickAndPlace evaluation results as well (Figure \ref{fig:fetch-generation}).

\end{document}